\documentclass{article}

\usepackage{microtype}
\usepackage{graphicx}
\usepackage{booktabs} 
\usepackage{eso-pic}
\usepackage{hyperref}

\usepackage[accepted]{icml2025}

\usepackage{amsmath}
\usepackage{amssymb}
\usepackage{mathtools}
\usepackage{amsthm}

\usepackage[capitalize,noabbrev]{cleveref}

\usepackage{minitoc}
\usepackage{CJKutf8} 
\usepackage{wrapfig}
\usepackage{lipsum} 
\usepackage{wrapfig}
\usepackage{subcaption}
\usepackage{hyperref}
\usepackage{url}
\usepackage{multicol}
\usepackage{aliascnt}
\usepackage{adjustbox}
\usepackage{tcolorbox}
\usepackage{booktabs}
\usepackage{multirow} 
\usepackage{todonotes}  
\usepackage{enumitem}
\usepackage{float}
\usepackage{multirow}
\usepackage[table]{xcolor} 
\usepackage{booktabs}
\usepackage{algorithm}
\usepackage{graphicx} 
\usepackage{subcaption} 
\usepackage{siunitx}
\usepackage{layout} 

\theoremstyle{plain}

\theoremstyle{definition}

\theoremstyle{remark}

\icmltitlerunning{RLoop: An Self-Improving Framework for Reinforcement Learning with Iterative Policy Initialization}

\begin{document}

% ...

\AddToShipoutPictureBG*{%
  \AtPageUpperLeft{%
    \hspace{1.5cm}%
    \raisebox{-3cm}{%
      \includegraphics[width=2cm]{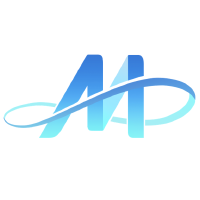}%
    }%
  }
  \AtPageUpperLeft{%
    \hspace{\paperwidth}
    \hspace{-3cm}
    \raisebox{-2.5cm}{%
      \rlap{\includegraphics[width=1cm]{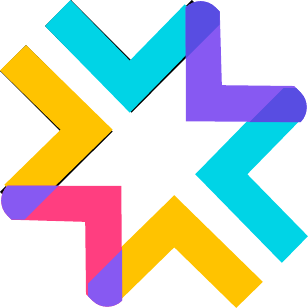}}%
    }%
  }
}



\twocolumn[
\icmltitle{RLoop: An Self-Improving Framework for Reinforcement Learning with Iterative Policy Initialization}



\icmlsetsymbol{equal}{*}

\begin{icmlauthorlist}
\icmlauthor{Zhiyuan Zeng}{fudan,seed}
\icmlauthor{Jiashuo Liu}{seed}
\icmlauthor{Zhangyue Yin}{fudan}
\icmlauthor{Ge Zhang\textsuperscript{\textdagger}}{seed}
\icmlauthor{Wenhao Huang\textsuperscript{\textdagger}}{seed}
\icmlauthor{Xipeng Qiu\textsuperscript{\textdagger}}{fudan,SII}
\end{icmlauthorlist}

\icmlaffiliation{fudan}{Fudan University}
\icmlaffiliation{seed}{Bytedance, seed}
\icmlaffiliation{SII}{Shanghai Innovation Institute}
\centering \textsuperscript{1}Fudan University \quad\quad\quad \textsuperscript{2}M-A-P \quad\quad\quad 
\textsuperscript{3}Shanghai Innovation Institute \\
\centering cengzy23@m.fudan.edu.cn \quad gezhang@umich.edu \quad rubio8741@gmail.com \quad xpqiu@fudan.edu.cn
\icmlkeywords{Machine Learning, ICML}

\vskip 0.3in
]

\renewcommand{\thefootnote}{} 
\footnotetext{\textdagger Corresponding authors}
\renewcommand{\thefootnote}{} 



\begin{abstract}
While Reinforcement Learning for Verifiable Rewards (RLVR) is powerful for training large reasoning models, its training dynamics harbor a critical challenge: ``RL overfitting,'' where models gain training rewards but lose generalization. Our analysis reveals this is driven by policy over-specialization and catastrophic forgetting of diverse solutions generated during training. Standard optimization discards this valuable inter-step policy diversity. To address this, we introduce \textbf{RLoop}, a self-improving framework built on iterative policy initialization. RLoop transforms the standard training process into a virtuous cycle: it first uses RL to explore the solution space from a given policy, then filters the successful trajectories to create an expert dataset. This dataset is used via Rejection-sampling Fine-Tuning (RFT) to refine the initial policy, creating a superior starting point for the next iteration. This loop of exploration and exploitation via iterative re-initialization effectively converts transient policy variations into robust performance gains. Our experiments show RLoop mitigates forgetting and substantially improves generalization, boosting average accuracy by 9\% and pass@32 by over 15\% compared to vanilla RL.
\end{abstract}

\section{Introduction}
\begin{figure*}[t]
    \centering
    \begin{subfigure}[b]{0.3\textwidth}
        \centering
        \includegraphics[width=\textwidth]{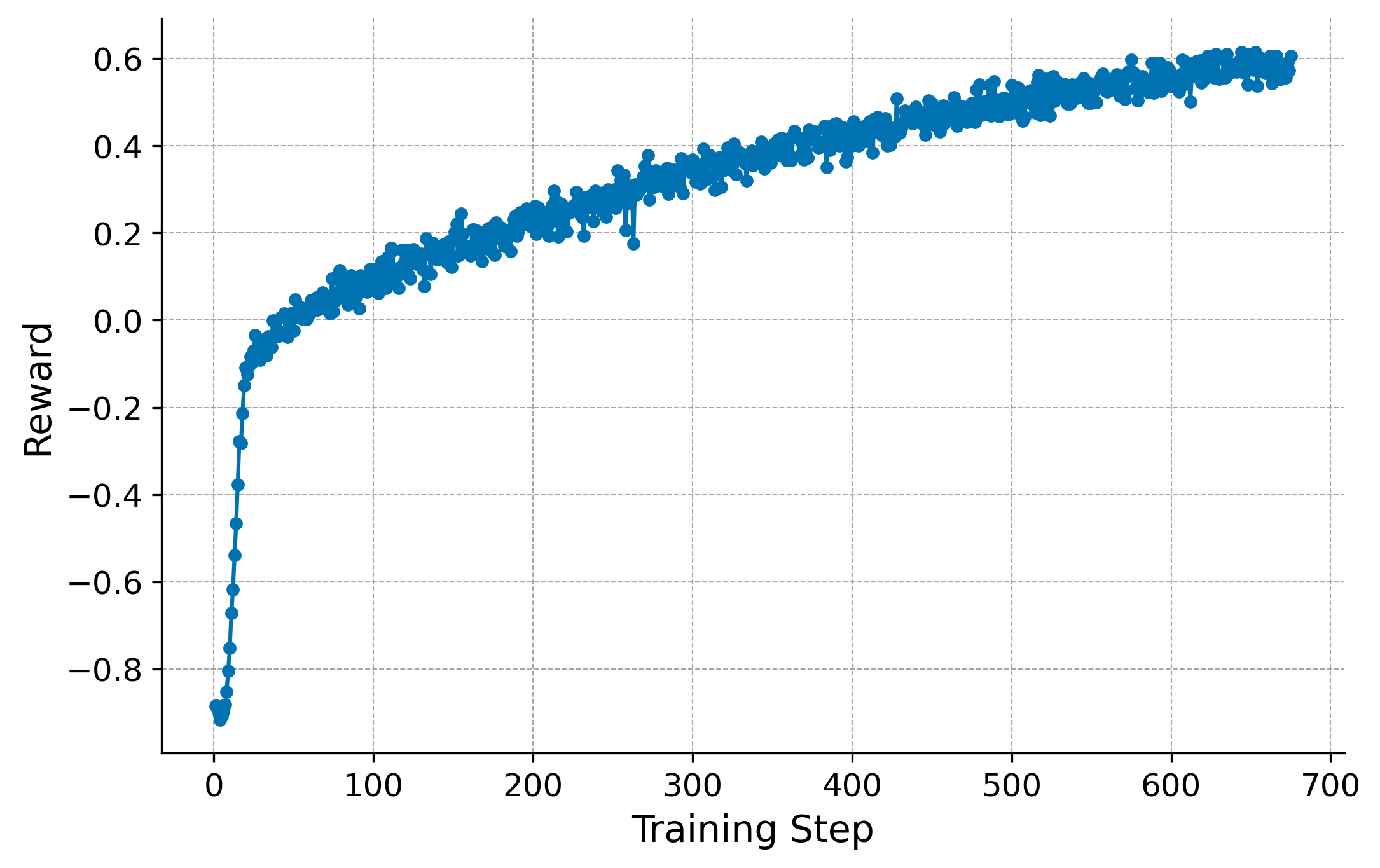}
        \caption{Reward on Training set.}
        \label{fig:reward}
    \end{subfigure}
    \hfill 
    \begin{subfigure}[b]{0.3\textwidth}
        \centering
        \includegraphics[width=\textwidth]{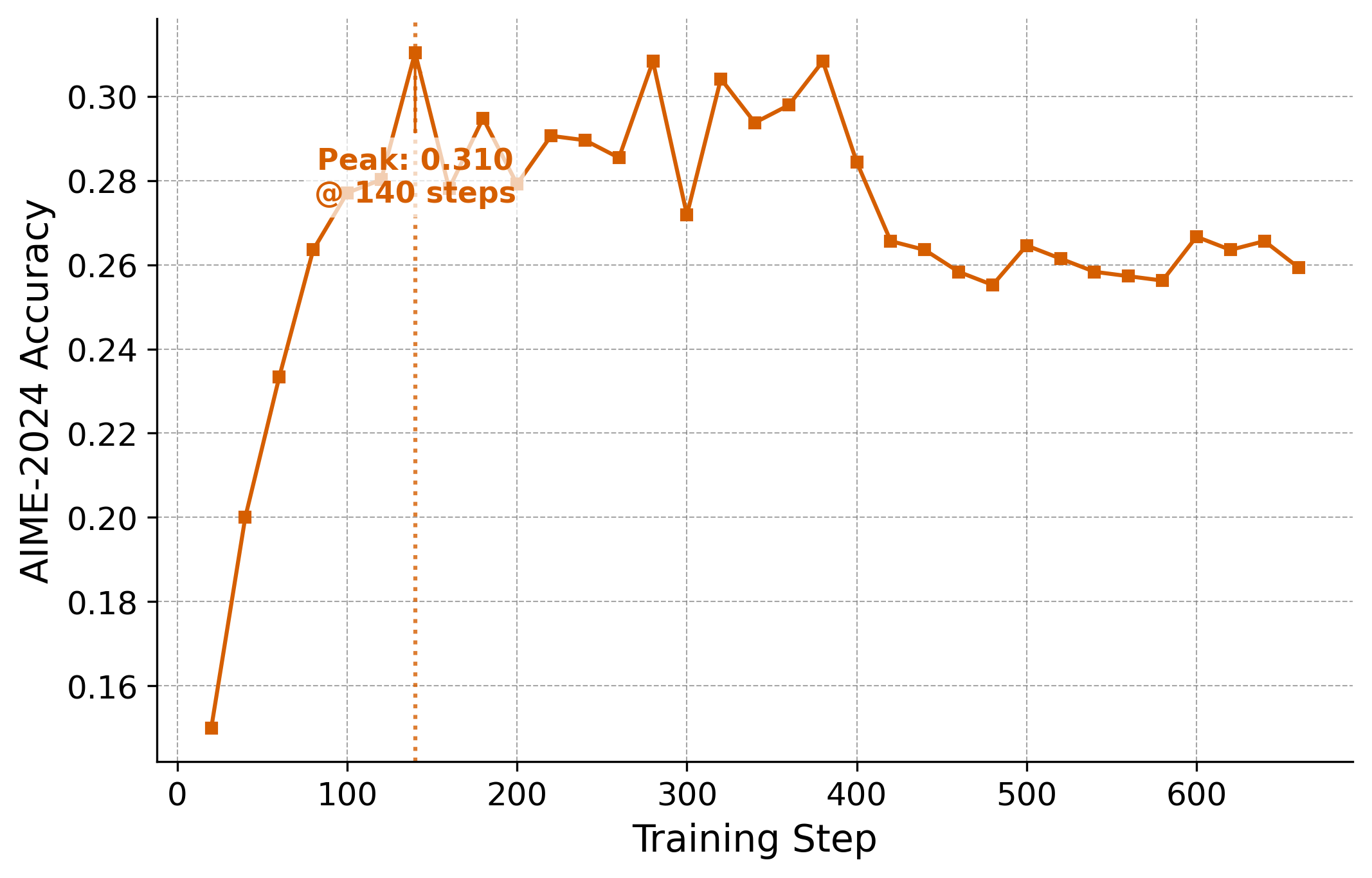}
        \caption{Accuracy on validation set.}
        \label{fig:acc}
    \end{subfigure}
    \hfill 
    \begin{subfigure}[b]{0.3\textwidth}
        \centering
        \includegraphics[width=\textwidth]{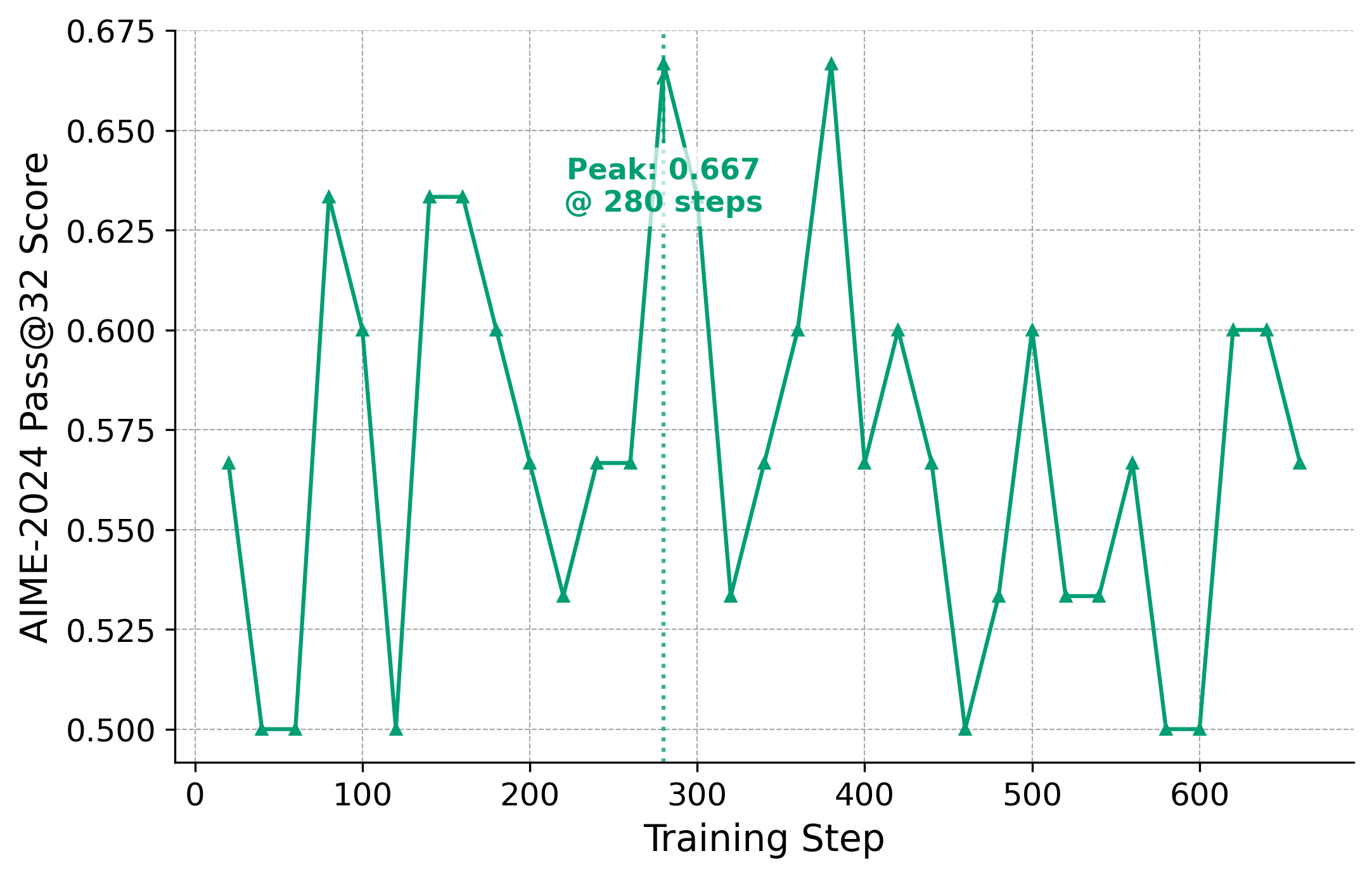}
        \caption{Pass@32 on validation set. }
        \label{fig:pass_at_32}
    \end{subfigure}
    
    \caption{The reward, accuracy and pass@32 score of Qwen-2.5-math-7b trained with the DAPO algorithm evaluated on AIME-2024.}
    \label{fig:train-acc-pass-at-k} 
\end{figure*}

Reinforcement Learning (RL), particularly through policy gradient methods such as PPO and its variants, has emerged as a cornerstone for aligning Large Language Models (LLMs) with complex human objectives. By enabling optimization for non-differentiable reward signals, RL has catalyzed significant advancements in diverse domains, including instruction following \cite{instruct-gpt} and mathematical reasoning \cite{deepseek-r1}.

However, our investigation into the application of RL for complex reasoning tasks reveals a critical, yet previously under-explored, challenge: a phenomenon we term \textbf{RL overfitting}, analogous to its supervised learning counterpart. As illustrated in Figure \ref{fig:train-acc-pass-at-k}, we observe a stark divergence between the training objective and true generalization performance. While the in-distribution reward signal exhibits a steady increase throughout training (e.g., for over 700 steps), the model's generalization capabilities—measured by out-of-distribution test accuracy and pass@k metrics—stagnate or even degrade much earlier (e.g., around 140 steps). This divergence strongly suggests that the RL agent becomes overly specialized in exploiting high-reward trajectories within its known distribution, leading to a model that is confident yet brittle when confronted with unseen problems.

To dissect the underlying dynamics of this overfitting phenomenon, we conducted a deeper empirical analysis of the RL training process. As shown in Figure \ref{fig:forgetting_matrix}, we found that standard RL training suffers from \textbf{catastrophic forgetting}, especially in the later stages, where the model discards approximately 30\% of the knowledge acquired during early training. This finding indicates that policies at different training steps are substantially distinct. Such inter-step policy diversity represents a valuable asset for exploration, yet it is typically discarded in conventional training paradigms. While prior work has acknowledged the importance of trajectory diversity \citep{entropy-rl,beyond-passat2}, it has primarily focused on the diversity generated by a single policy at a fixed step, overlooking the rich diversity across different training checkpoints. 

Inspired by the potential of harnessing this inter-step diversity, we propose \textbf{RLoop}, a self-improving framework centered around iterative policy initialization. Instead of a single, monolithic training run, RLoop recasts the process as a virtuous cycle where the policy is progressively refined and re-initialized. Each iteration consists of two phases:
\begin{enumerate}
    \item \textbf{Exploration Phase:} Starting from the current policy $\theta_i$, we run a standard RL process not to find a single optimal policy, but to generate a diverse pool of solution trajectories. The significant policy shifts across RL steps act as a powerful, built-in exploration mechanism.
    \item \textbf{Exploitation Phase:} We curate the trajectories generated during exploration by filtering for successful outcomes. This ``expert'' dataset, $D_{\text{expert}}^i$, is then used to refine the initial policy $\theta_i$ via Supervised Fine-Tuning. The resulting improved policy, $\theta_{i+1}$, serves as a superior starting point for the subsequent exploration phase.
\end{enumerate}
The crucial step is that this consolidated policy $\theta_{i+1}$ is not the final output, but serves as a superior initial policy for the next Exploration Phase. RLoop thus establishes a self-contained improvement loop: RL explores possibilities from a stable base, and RFT consolidates the findings into a better base. This process of iterative policy initialization allows the model to systematically accumulate knowledge, turning the transient diversity from RL into robust, generalizable capabilities. Unlike prior works that rely on external expert data to bridge RL and supervised learning \cite{luffy,chord,bridge}, RLoop bootstraps its own progress. To further stabilize this self-improvement, we incorporate an active learning strategy to ensure the model continually focuses on challenging problems.

Our main contributions are summarized as follows:
\begin{itemize}
    \item We empirically identify and characterize the ``RL overfitting'' phenomenon in LLMs, demonstrating that reward improvements do not necessarily translate to enhanced generalization.
    \item We reveal that this overfitting is linked to catastrophic forgetting and highlight the untapped potential of inter-step policy diversity, a valuable resource discarded by standard RL.
    \item We propose \textbf{RLoop}, an iterative self-improvement framework that effectively balances exploration and exploitation by alternating between RL for diverse solution generation and RFT for knowledge consolidation.
    \item Our experiments demonstrate that RLoop significantly outperforms vanilla RL on challenging math reasoning benchmarks, particularly in pass@k metrics. We further show that RLoop mitigates forgetting and make the RL training more stable.
\end{itemize}
\section{Related Works}
\paragraph{Generalization of RLVR.}
The generalization capabilities of RLVR are a subject of active research with conflicting findings. Several theoretical and empirical studies suggest that RL can lead to strong generalization \citep{sft-rl-generalizes, med-rlvr, causal-rl}, with some work even demonstrating its effectiveness with a single question-answer pair \citep{one-shot-rl}. However, this optimistic view is challenged by other research. \citet{limited-rl} empirically found that while RLVR improves greedy-decoding accuracy, it can degrade pass@k performance, especially for large $k$. This suggests that standard RL may merely improve test-time efficiency rather than enhancing the model's core ability to solve novel problems. Conversely, other studies \citep{prolong-rl} report that RLVR can indeed improve pass@k scores on certain tasks. Our work contributes to this debate by identifying a key dynamic: both accuracy and pass@k can degrade during training due to an overfitting-like phenomenon, which we aim to resolve.

Efforts to improve RLVR generalization can be categorized into three main perspectives:
\begin{enumerate}
    \item \textbf{Data-centric Approaches:} These methods focus on enriching the training data. For instance, \citet{QuestA} and \citet{beyond-passat1} propose augmenting the question set to expose the model to a wider range of states. Others focus on curriculum learning; \citet{zero-reward} found that mixing simple and hard questions facilitates knowledge transfer, while \citet{knapsack-rl} advocate for increasing the rollout budget for more challenging problems, based on the finding that performance correlates with the number of unique problems solved.
    \item \textbf{Algorithm-centric Approaches:} These methods modify the RL algorithm itself. A prominent line of work explores the relationship between performance and policy entropy \citep{entropy-rl, beyond-passat2}. \citet{beyond-passat2} propose masking gradients from low-entropy tokens to focus learning on more uncertain parts of the reasoning process. Similarly, \citet{entropy-rl} introduce methods like Clip-Cov and KL-Cov to constrain gradients from high-variance tokens, thereby enhancing exploration.
    \item \textbf{Initialization-centric Approaches:} Recognizing that a strong starting point is crucial for RL, these approaches focus on creating a superior initial policy. Works like \citet{OctoThinker} and \citet{openthoughts} achieve this by pre-training or fine-tuning models on large, high-quality, reasoning-intensive corpora before applying RL.
\end{enumerate}

Our proposed RLoop framework aligns with the initialization-centric perspective but with a critical distinction: it operates as a \textbf{self-improving}, iterative re-initialization loop. Unlike methods requiring extensive human effort for data curation, RLoop synthesizes its own "expert" data for re-initialization directly from the trajectories generated during the RL phase, creating a fully autonomous improvement cycle.

\paragraph{Combining RL and SFT.}
The RLoop framework, which iterates between RL and Rejection-sampling Fine-Tuning (RFT, a form of SFT), belongs to a broader class of methods that combine SFT and RL. The most common paradigm is a simple pipeline where SFT provides the initial policy for a subsequent RL phase \citep{instruct-gpt}.

More intricate integrations have also been explored. Some methods aim to incorporate SFT data directly into the RL process, for instance, through off-policy learning \citep{luffy}. Others attempt to merge the SFT and RL objectives into a single loss function for joint training \citep{chord, luffy, bridge}. A different approach, exemplified by \citet{ReLIFT}, involves interleaving SFT steps within the RL training loop, using high-quality solutions discovered during RL to reinforce the policy. Addressing the catastrophic forgetting issues observed in such methods, \citet{MIFO} proposed constraining the SFT updates. Another line of work, including \citet{self-rewarding} and \citet{BRiTE}, formulates the problem as a latent variable model akin to a Variational Autoencoder (VAE) \citep{vae}, where SFT updates a reward model and RL improves the policy in an alternating fashion.

Our RLoop framework is distinct from these prior works in two fundamental ways. First, it employs a cyclical, macro-level iteration between distinct RFT and RL phases, rather than a fine-grained interleaving or joint loss. Second, and more importantly, RLoop is entirely self-contained, bootstrapping its SFT data from the RL agent's own successful explorations. This eliminates the need for any external expert data, setting it apart from most hybrid SFT-RL methods.

\section{Preliminary Study: Characterizing Policy Dynamics in RLVR}
\label{sec:prelimary-study}
As established in the introduction and illustrated in Figure \ref{fig:train-acc-pass-at-k}, standard RLVR exhibits an overfitting-like behavior, where training rewards diverge from validation performance. To delve into the mechanisms behind this phenomenon, we conduct a preliminary study analyzing three key metrics: the learning rate, the forgetting rate, and trajectory similarity. The learning and forgetting rates quantify the model's ability to acquire new problem-solving capabilities and its tendency to lose previously acquired ones, respectively. Trajectory similarity measures the distributional shift in generated solutions across different training steps.

\begin{figure*}[t]
    \centering
    
    \begin{subfigure}[b]{0.32\linewidth}
        \centering
        \includegraphics[width=\linewidth]{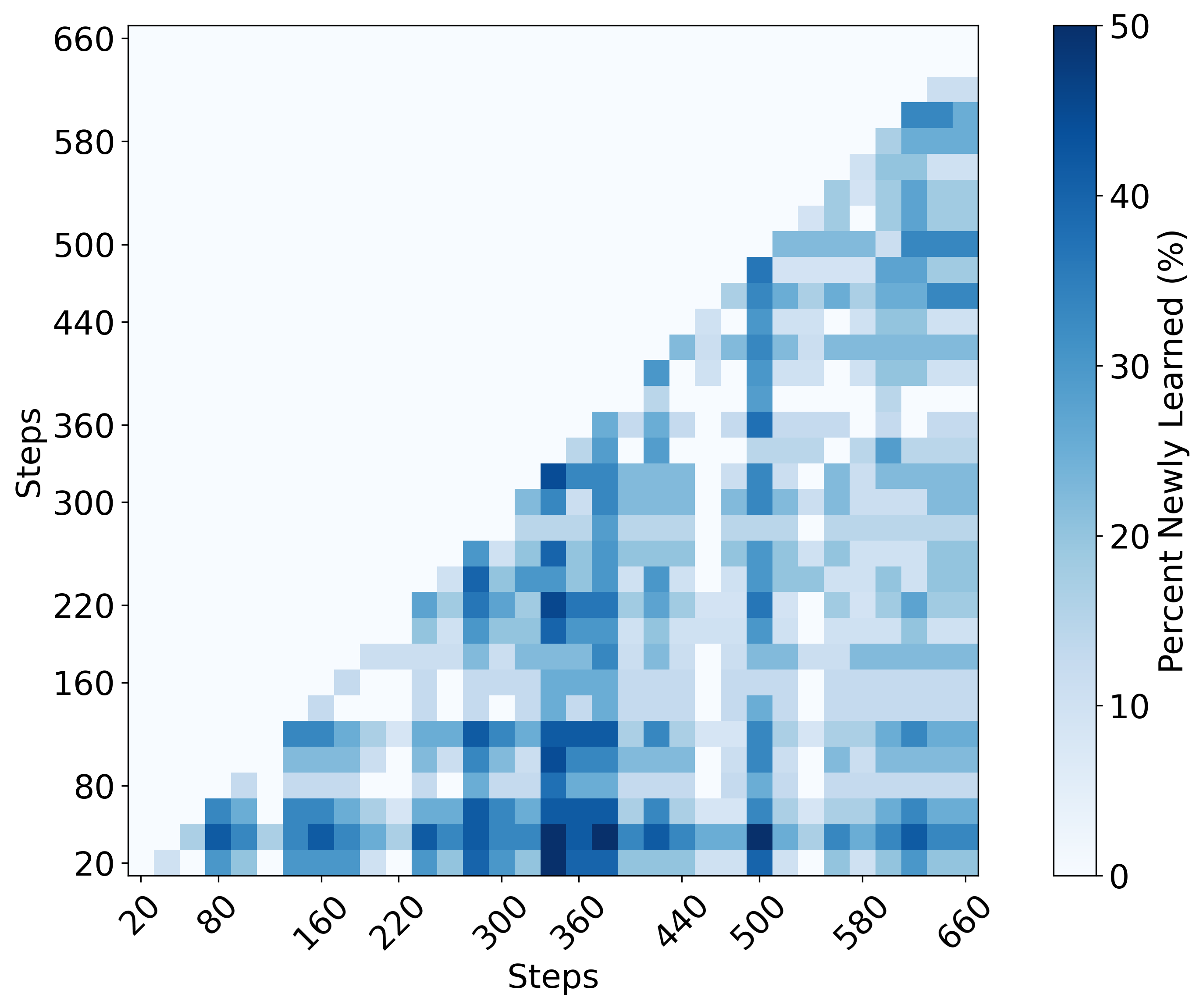}
        \caption{Learning Matrix}
        \label{fig:learning_matrix}
    \end{subfigure}
    \hfill
    \begin{subfigure}[b]{0.32\linewidth}
        \centering
        \includegraphics[width=\linewidth]{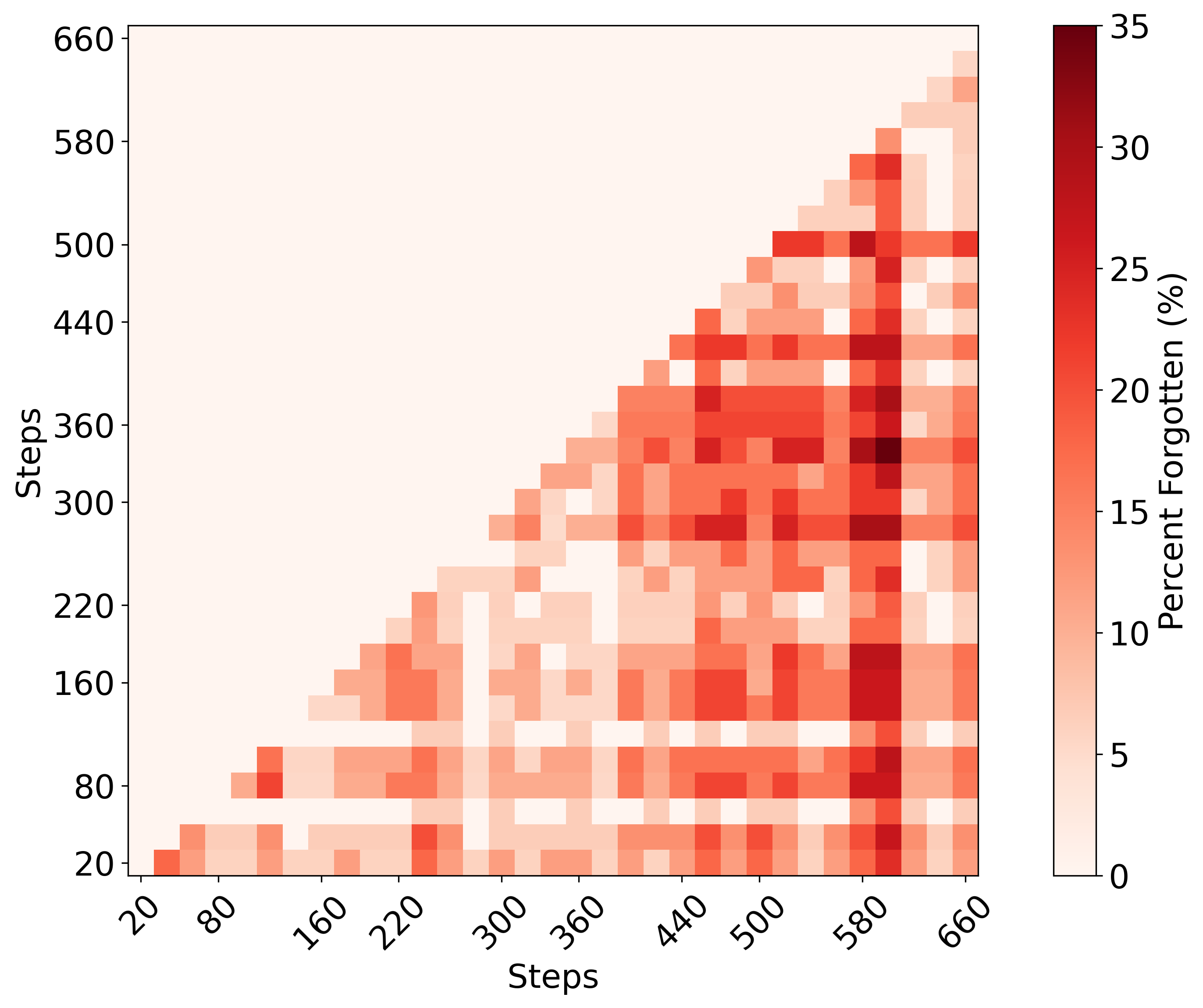}
        \caption{Forgetting Matrix}
        \label{fig:forgetting_matrix}
    \end{subfigure}
    \hfill
    \begin{subfigure}[b]{0.32\linewidth}
        \centering
        \includegraphics[width=\linewidth]{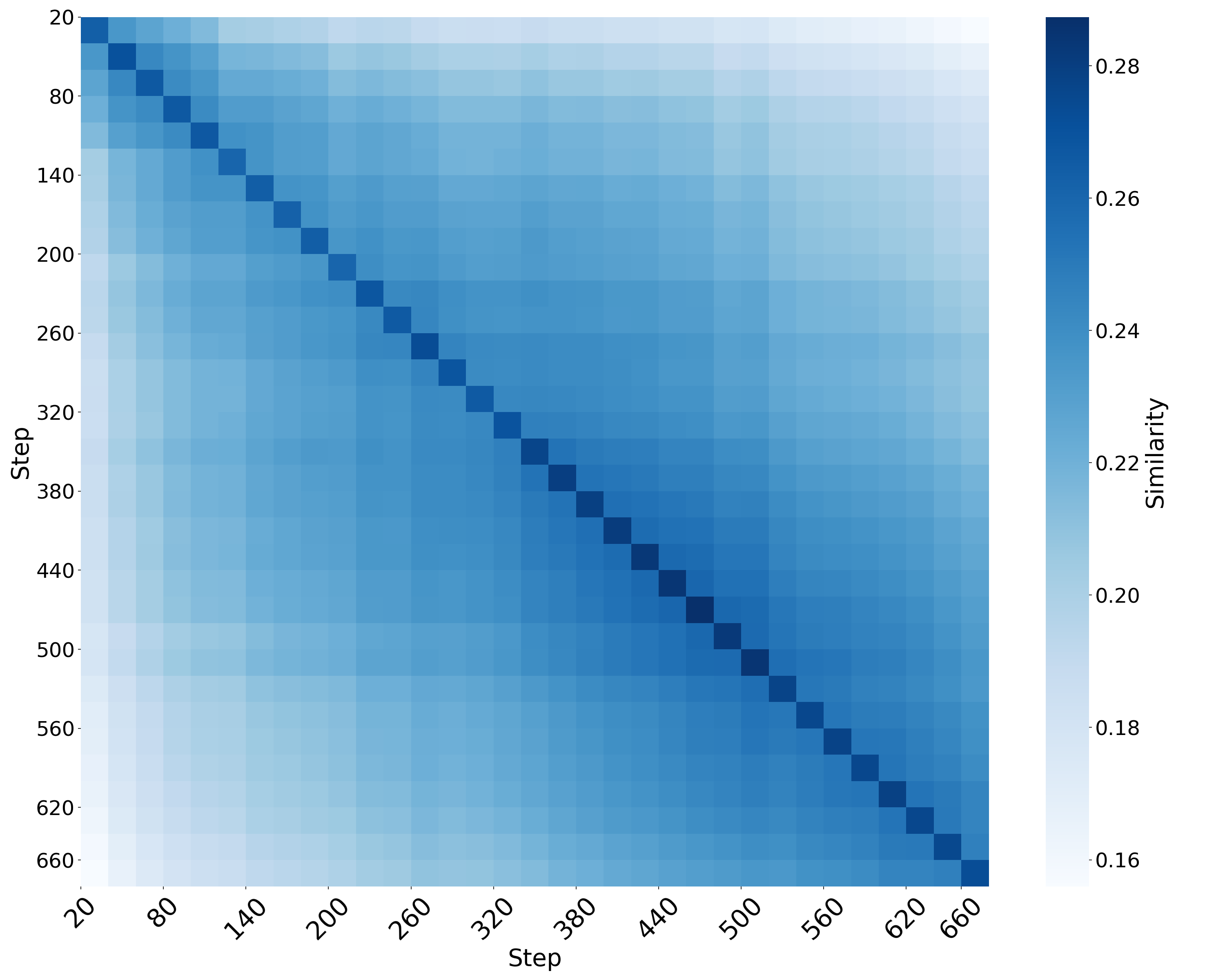}
        \caption{Similarity Matrix}
        \label{fig:similarity_matrix}
    \end{subfigure}
    \caption{The value at ($i$, $j$) in the Learning Matrix represents the percentage of validation problems that the policy at step $j$ can solve but the policy at step $i$ cannot. Conversely, the value in the Forgetting Matrix represents problems solvable at step $i$ but not at step $j$. The value at ($i$, $j$) in the Similarity Matrix indicates the average n-gram similarity of trajectories between policies from step $i$ and step $j$. For all analyses, we sample 32 solutions for each question in the validation set.}
    \label{fig:diff-across-steps}
\end{figure*}

\subsection{Learning and Forgetting Dynamics}
To understand the trade-offs during training, we analyze the policy's ability to both learn new problems and forget old ones. We define the learning rate from a checkpoint at step $i$ to a later checkpoint at step $j$ ($i < j$) as the proportion of validation problems that the policy at step $j$ can solve, but the policy at step $i$ could not. Symmetrically, the forgetting rate is the proportion of problems that the policy at step $i$ could solve, but the policy at step $j$ fails to solve.

The Learning Matrix (Figure \ref{fig:learning_matrix}) reveals that the model continuously learns to solve new problems throughout training. The non-zero values in the upper triangle indicate that later policies acquire capabilities that earlier policies lacked. This demonstrates that continued training is not merely fitting to noise; the agent is genuinely expanding its problem-solving repertoire.

However, this learning comes at a cost, as shown by the Forgetting Matrix (Figure \ref{fig:forgetting_matrix}). The matrix reveals a significant level of forgetting, with rates frequently exceeding 10\% and reaching as high as 35\%, particularly between distant checkpoints. This observation empirically confirms that the RLVR is not only acquiring new skills but is also simultaneously discarding previously learned ones.

The interplay between learning and forgetting explains the performance stagnation observed in RLVR. In the early stages of training, the learning rate surpasses the forgetting rate, leading to a net increase in validation performance. As training progresses, the forgetting rate begins to catch up with or even exceed the learning rate. Therefore, the model's performance on the validation set oscillates or decreases. This dynamic provides a compelling explanation for why prolonged training does not necessarily lead to better generalization.

\subsection{Similarity Matrix}
To complement the performance-based analysis of the forgetting matrix, we analyze the lexical similarity of the generated trajectories. This metric quantifies the textual consistency of solutions generated by policies at two distinct training steps, $i$ and $j$.

The core idea is to measure the overlap of n-grams (specifically, bi-grams) between the sets of solutions generated at these two steps. The process involves calculating the pairwise similarity for all solution pairs using the Jaccard index, and then aggregating these scores into a single, robust metric. The detailed mathematical formulation for this calculation is deferred to Appendix \ref{app:similarity-calculation}.

The resulting Similarity Matrix (Figure \ref{fig:similarity_matrix}) shows that the intra-step similarity (diagonal entries, typically $>$0.26) is substantially higher than the inter-step similarity (off-diagonal entries, typically $<$0.2). Moreover, the similarity systematically decreases as the distance between steps increases.

Taken together, the Learning, Forgetting, and Similarity Matrices provide compelling evidence from two different perspectives—performance and solution form—that policies at different RL training steps are remarkably diverse. This motivates our core idea: to explicitly collect and consolidate this valuable, yet typically discarded, diversity for more robust generalization.

\section{Methodology}
\label{sec:method}

\begin{algorithm*}[t]
\caption{Iterative Policy Initialization}
\label{alg:reft-rl}
\begin{algorithmic}
   \STATE {\bfseries Initialize:} Start with a base policy $\pi_{\theta_0}$ (e.g., Qwen-2.5-7b-math).
   \FOR{$i=0$ {\bfseries to} $I-1$}
      \STATE \textit{// --- Exploration Step ---}
      \STATE Initialize RL policy from $\pi_{\theta_i}$.
      \STATE Run RL for $N_{RL}$ steps to generate a set of trajectories $D_{\text{RL}}^i$.
      \STATE \textit{// --- Exploitation Step ---}
      \STATE Filter for successful trajectories: $D_{\text{expert}}^i = \{\tau \in D_{\text{RL}}^i \mid R(\tau) > 0\}$.
      \STATE Initialize a new policy from the same starting point $\pi_{\theta_i}$.
      \STATE Perform Supervised Fine-Tuning on this policy using $D_{\text{expert}}^i$ to obtain $\pi_{\theta_{i+1}}$.
      \STATE $\theta_{i+1} = \arg\max_{\theta} \sum_{\tau \in D_{\text{expert}}^i} \log \pi_\theta(\tau)$
   \ENDFOR
   \STATE {\bfseries Return:} The final refined policy $\pi_{\theta_I}$.
\end{algorithmic}
\end{algorithm*}
\subsection{RLoop Framework}
To counteract the overfitting and instability identified in Figure \ref{fig:train-acc-pass-at-k}, we introduce RLoop, an iterative training framework designed to harness the inter-step policy diversity. RLoop transforms the standard linear training process into a cyclical one, explicitly alternating between an RL-based exploration phase and an RFT-based exploitation phase.

\label{sec:framework}
The framework operates as an iterative loop, as detailed in Algorithm \ref{alg:reft-rl}:

\begin{enumerate}
    \item \textbf{Exploration Phase (RL):} Starting from a base policy $\pi_{\theta_i}$, we execute a standard RL training process for a fixed number of steps. The primary goal of this phase is not to train the policy to convergence, but to leverage it as a powerful search algorithm. The inherent stochasticity and policy drift across steps drive the model to explore diverse modes of the solution space. We collect trajectories from multiple intermediate checkpoints within this phase to create a rich and varied dataset, $D_{\text{RL}}^i = \{\tau_1, \dots, \tau_N\}$.

    \item \textbf{Exploitation Phase (RFT):} In this phase, we distill and consolidate the knowledge discovered during exploration. First, we filter the collected trajectories using the reward signal, retaining only successful solutions to form an ``expert'' dataset: $D_{\text{expert}}^i = \{\tau \in D_{\text{RL}}^i \mid R(\tau) > 0\}$. We then use this curated dataset to fine-tune the initial policy $\pi_{\theta_i}$ via Supervised Fine-Tuning (SFT). The resulting improved policy, $\pi_{\theta_{i+1}}$, becomes the starting point for the next iteration of the loop, thus creating a self-improving cycle.
\end{enumerate}

 The overall process is summarized in Algorithm \ref{alg:reft-rl}.

\subsection{Active Learning for Focused Exploitation}
\label{ssec:dynamics_and_active_learning}

The efficacy of the RLoop framework arises from the complementary strengths of RL and RFT. A closer look at their optimization dynamics reveals why RFT provides stable exploitation and, crucially, why this stability benefits from an active learning strategy.

The policy gradients for RL (e.g., REINFORCE) and RFT differ fundamentally in their weighting schemes:
\begin{align}
    \nabla_\theta J_{\text{RL}}(\theta) &= \mathbb{E}_{\tau \sim \pi_\theta} [A(\tau) \nabla_\theta \log \pi_\theta(\tau)] \\
    \nabla_\theta J_{\text{RFT}}(\theta) &= \mathbb{E}_{\tau \sim \pi_{\theta_{\text{RL}}}} [R(\tau) \nabla_\theta \log \pi_\theta(\tau)]
\end{align}
\begin{itemize}
    \item \textbf{RL's Relative Weighting:} The advantage function $A(\tau)$ measures a trajectory's quality \textit{relative} to the policy's average performance. This makes RL effective at differentiating good from better but provides a vanishing learning signal when all sampled trajectories are already successful (i.e., their advantages are near zero). 

    \item \textbf{RFT's Absolute Weighting:} In contrast, RFT uses the \textit{absolute} reward $R(\tau)$ as a weight (effectively 1 for success and 0 for failure). This provides a stable, low-variance learning signal that reinforces every successful trajectory, regardless of the batch's overall performance.
\end{itemize}

While RFT's stability is a key advantage, it introduces a potential inefficiency: RFT may over-invest capacity on problems the model already solves consistently. Since every successful trajectory receives an equal weight of 1, the model might spend excessive effort reinforcing its knowledge of ``easy'' problems.

This observation directly motivates our use of active learning. To make the exploitation phase more efficient and targeted, we apply a filter before the RFT step. We identify a subset of problems that the current policy finds ``hard'' (e.g., those with a low success rate across generated samples). The RFT update is then performed exclusively on successful trajectories from this hard subset. This active learning strategy ensures that the model's capacity is focused on expanding its capabilities at the frontier of its knowledge, preventing redundant updates on mastered tasks and optimizing computational resource usage.

\begin{table*}[h]
\centering
\caption{The performance comparison between the base model (Qwen-2.5-7b-Math), RL and RLoop.}
\label{tab:main-results}
\begin{tabular}{llcccc}
\toprule
\textbf{Dataset} & \textbf{Method} & \textbf{Avg@32} & \textbf{Pass@8} & \textbf{Pass@16} & \textbf{Pass@32} \\
\midrule
\multirow{3}{*}{AIME} & \textit{Base}& 6.00 & 43.65 & 46.09 & 46.66 \\
& \textit{RL} & 31.04 & 50.96 & 56.97 & 63.33 \\
& \textit{RLoop} & \textbf{37.60} & \textbf{58.77} & \textbf{66.69} & \textbf{73.33} \\
\midrule
\multirow{3}{*}{MinervaMath} & \textit{Base}& 8.49 & 25.78 & 29.36 & 31.61 \\
& \textit{RL} & 18.37 & 27.64 & 29.27 & 29.63 \\
& \textit{RLoop} & \textbf{19.88} & \textbf{29.95} & \textbf{31.58} & \textbf{32.59} \\
\midrule
\multirow{3}{*}{Omini-Math} & \textit{Base}& 8.00 & 26.38 & 31.58 & 36.20 \\
& \textit{RL} & 19.81 & 27.70 & 29.23 & 30.00 \\
& \textit{RLoop} & \textbf{21.00} & \textbf{31.74} & \textbf{34.58} & \textbf{37.00} \\
\midrule
\multirow{3}{*}{MATH} & \textit{Base}& 24.71 & 67.32 & 73.19 & 76.00 \\
& \textit{RL} & 56.78 & 66.54 & 68.64 & 70.20 \\
& \textit{RLoop} & \textbf{58.93} & \textbf{75.50} & \textbf{78.08} & \textbf{80.00} \\
\midrule
\multirow{3}{*}{Avg} & \textit{Base} & 16.80 & 40.78 & 45.06 & 47.62 \\
& \textit{RL} & 31.50 & 43.21 & 46.03 & 48.29 \\
& \textit{RLoop} & \textbf{34.35} & \textbf{49.00} & \textbf{52.73} & \textbf{55.73} \\
\bottomrule
\end{tabular}
\end{table*}

\subsection{Theoretical Grounding: RFT as Importance-Weighted MLE}
The RFT phase is not merely a heuristic; it can be theoretically grounded as a form of policy improvement derived from Maximum Likelihood Estimation (MLE) with importance sampling.

Ideally, we would want to train our policy $\pi_\theta$ to match an unknown "expert" distribution $p^*(\tau)$ that produces correct and generalizable solutions. This corresponds to maximizing the log-likelihood:
\begin{equation}
    \mathcal{L}_{\text{MLE}}(\theta) = \mathbb{E}_{\tau \sim p^*}[\log \pi_\theta(\tau)].
\end{equation}
We cannot sample directly from $p^*$, but we can sample from the RL policy which can be seen as an approximation of $p^*$. Therefore, we use importance sampling to re-express this objective using trajectories sampled from our RL policy, $\pi_{\theta_{\text{RL}}}$:
\begin{equation}
    \mathcal{L}_{\text{MLE}}(\theta) = \mathbb{E}_{\tau \sim \pi_{\theta_{\text{RL}}}} \left[ \frac{p^*(\tau)}{\pi_{\theta_{\text{RL}}}(\tau)} \log \pi_\theta(\tau) \right].
    \label{eq:is}
\end{equation}
The key challenge is the unknown importance weight $w(\tau) = p^*(\tau) / \pi_{\theta_{\text{RL}}}(\tau)$. However, we can approximate this weight using the reward signal $R(\tau)$. Intuitively, a trajectory $\tau$ with a high reward is more likely to belong to the expert distribution $p^*$ than a trajectory with a low reward. For binary rewards ($R(\tau) \in \{0, 1\}$), this leads to a simple and powerful approximation:
\begin{equation}
    w(\tau) = \frac{p^*(\tau)}{\pi_{\theta_{\text{RL}}}(\tau)} \propto R(\tau).
\end{equation}
By substituting this approximation into the importance-sampled objective (Equation \ref{eq:is}), we arrive at the RFT objective:
\begin{equation}
    \mathcal{L}_{\text{RFT}}(\theta) = \mathbb{E}_{\tau \sim \pi_{\theta_{\text{RL}}}} \left[ R(\tau) \log \pi_\theta(\tau) \right].
    \label{eq:rft}
\end{equation}
This objective is precisely the SFT loss applied to the reject-sampling dataset $D_{\text{expert}}$, thus providing a principled justification for our method.

\section{Experiments}
\subsection{Experiment Setting}
\paragraph{Datasets and Evaluation.}
For training, we employ the DAPO-17k dataset \citep{DAPO}, which consists of 17,000 challenging mathematical problems. To ensure a comprehensive assessment of generalization, we evaluate our models on a suite of widely recognized benchmarks: AIME 2024, MinervaMath \citep{minervamath}, OmniMath \citep{omini-math}, and the MATH-500 test set \citep{math-dataset}.

\paragraph{RL Setup.}
Our base model for all reinforcement learning experiments is Qwen-2.5-7b-Math. In line with the approach of R1-Zero \citep{deepseek-r1}, we apply RL directly to this base model with rule-based reward. We employ the DAPO algorithm \citep{DAPO} with a group size of 16 and a maximum generation length of 2048 tokens. For checkpoint selection, we use AIME 2024 as our validation set, selecting the model that achieves the best performance for final evaluation.

\paragraph{RLoop Setup}
For our proposed RLoop framework, the RFT phase utilizes trajectories cached during the preceding RL exploration phase. We implement an active learning strategy by filtering this data, retaining only successful trajectories from ``hard'' problems—defined as those where the model's success rate during the RL phase was below 10\%. Each full iteration of the RLoop cycle consists of 200 RL training steps followed by one epoch of RFT on the curated dataset.

\begin{figure*}[t]
    \centering
    \includegraphics[width=1\linewidth]{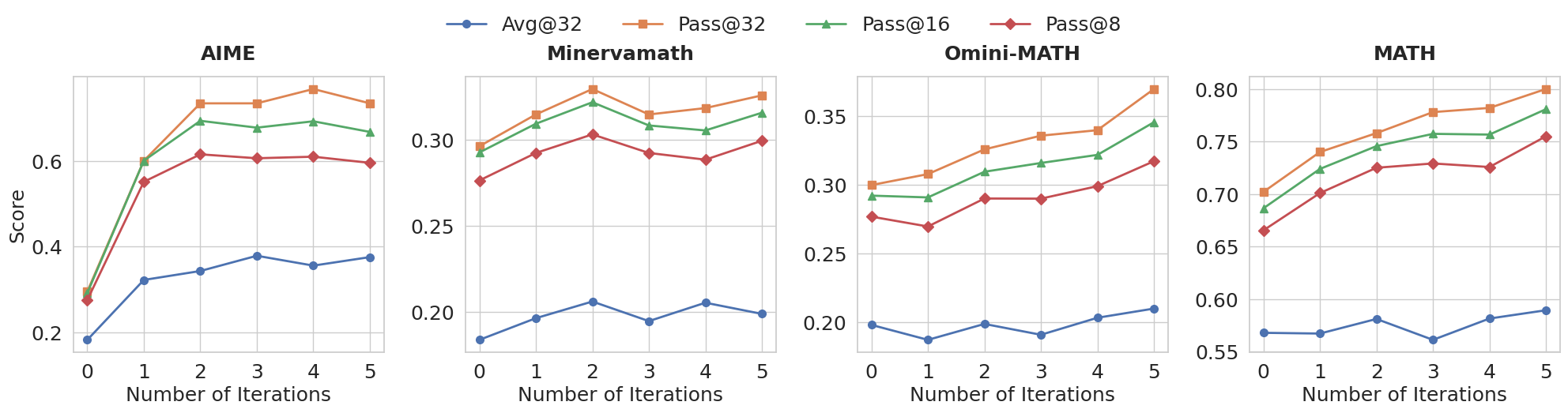}
    \caption{The performance of Qwen-2.5-7b-Math trained with RLoop in different number of iterations, in terms of accuracy and pass@k score.} 
    \label{fig:scale-reinit}
\end{figure*}

\begin{figure}[t] 
    \centering 
    \includegraphics[width=0.8\linewidth]{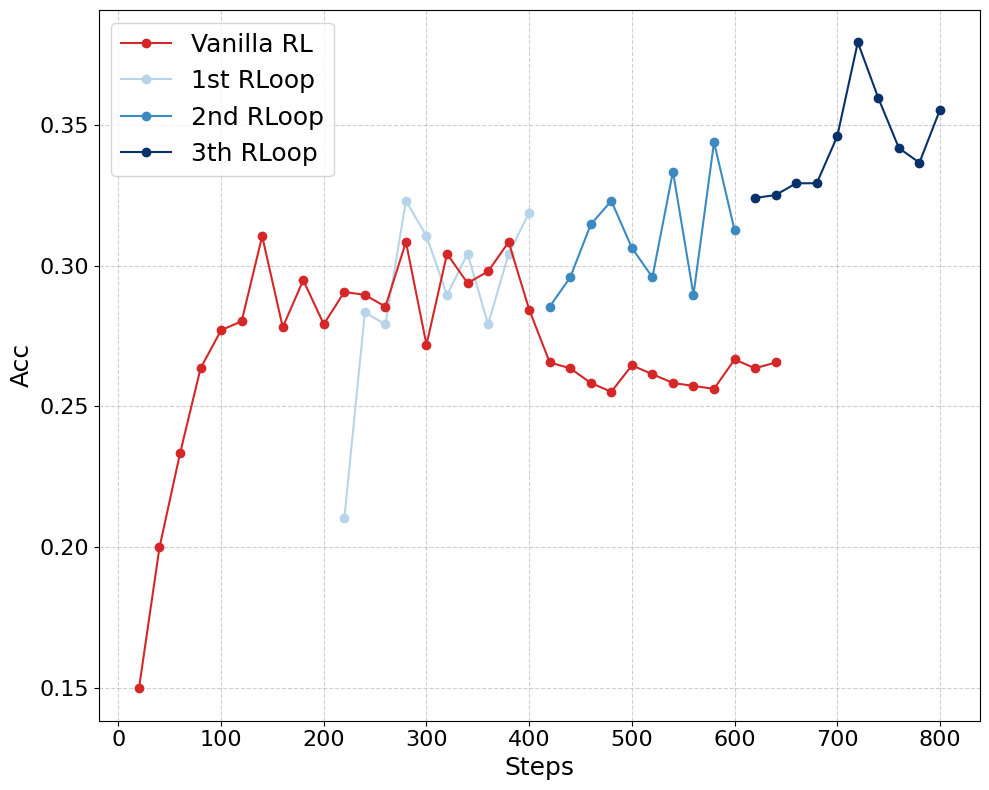}
    \caption{Compare the accuracy of vanilla RL an RLoop at different training steps.} 
    \label{fig:compare-reinit-rl-acc} 
\end{figure}

\subsection{Main Results}

We conduct a comprehensive comparison between our proposed RLoop framework and a standard vanilla RL baseline (DAPO). The results are presented in Table \ref{tab:main-results}. To ensure a fair comparison of computational costs, the vanilla RL baseline was trained for 600 steps, while RLoop was run for three iterations, with each iteration comprising 200 RL steps. The computational overhead of the RFT phase is negligible relative to the RL phase, making the total training budgets for both methods comparable.

As evidenced by the results, RLoop consistently and substantially outperforms the vanilla RL baseline across all evaluation benchmarks, in terms of both accuracy (Average@32) and Pass@k. The most significant gains are observed in the Pass@k scores, highlighting RLoop's ability to generate a more diverse set of correct solutions. As expected, the performance improvement on AIME 2024 is particularly pronounced, as it was used as the validation set for checkpoint selection.

A crucial observation is the detrimental effect of vanilla RL on the model's pass@k performance. On three out of four benchmarks (MinervaMath, Omni-Math, and MATH), the pass@k scores of the RL-trained model are worse than those of the original base model, especially at larger values of $k$ like $k=32$. This aligns with findings from prior work \citep{limited-rl}, which posited that standard RL might fail to genuinely enhance the intrinsic reasoning capabilities of LLMs. However, our RLoop framework not only reverses this degradation but surpasses the base model's performance by a significant margin. This suggests that the performance drop is not an inherent flaw of using RL for reasoning, but rather a byproduct of the standard, continuous training paradigm that leads to overfitting.

Interestingly, vanilla RL shows a performance gain on AIME 2024, in contrast to its degradation on other benchmarks. We hypothesize that this is due to a closer distributional similarity between the AIME 2024 dataset and the DAPO-17k training set. This further supports our claim that vanilla RL tends to overfit to the training distribution, leading to diminished performance on out-of-distribution (OOD) tasks. RLoop, by cyclically exploring and consolidating knowledge, effectively mitigates this issue and achieves superior generalization.

\begin{figure*}[t]
    \centering
    \begin{subfigure}[b]{0.32\linewidth}
        \centering
        \includegraphics[width=\linewidth]{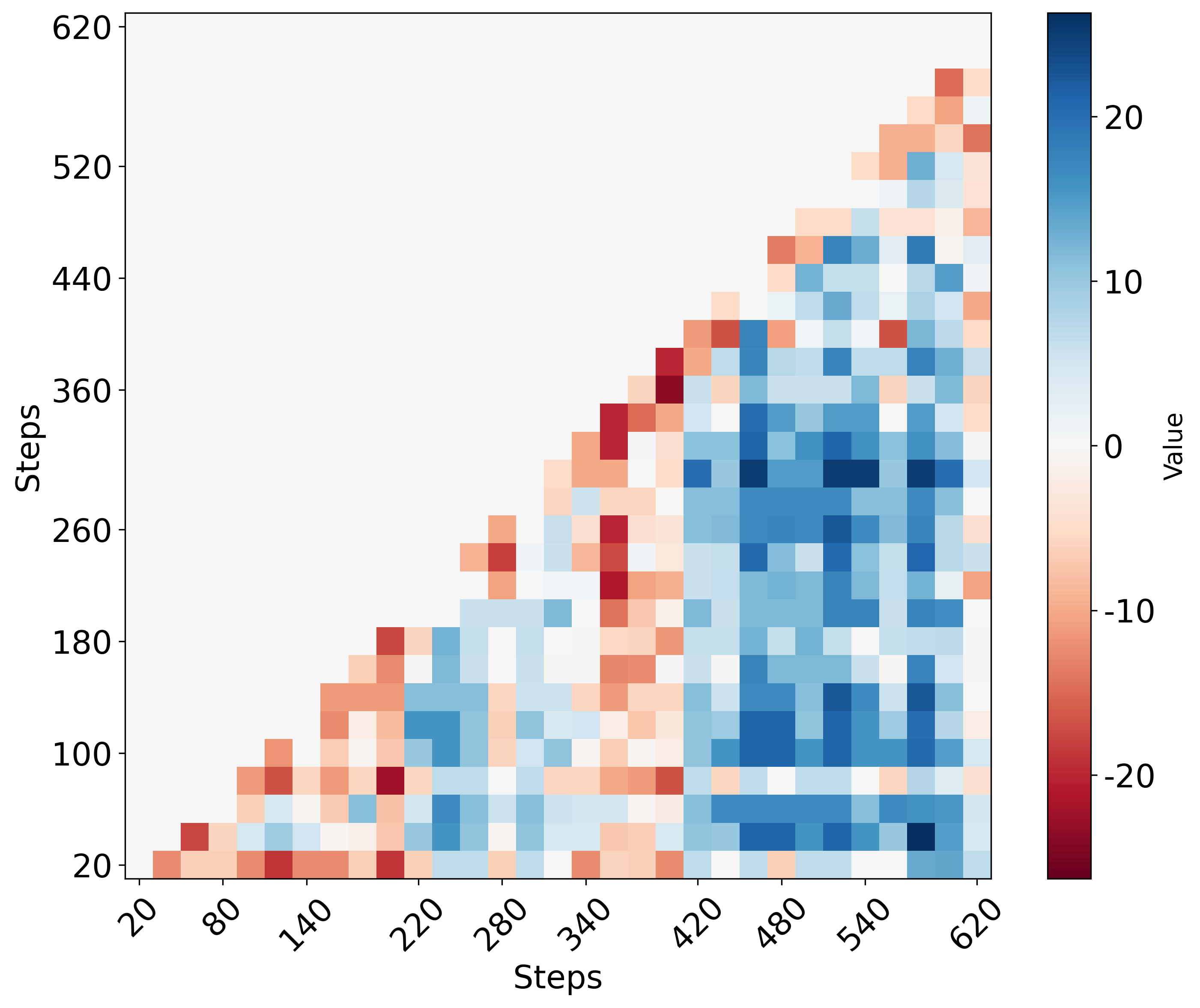}
        \caption{Forgetting Rate.}
        \label{fig:forget-comparison}
    \end{subfigure}
    \hfill
    \begin{subfigure}[b]{0.32\linewidth}
        \centering
        \includegraphics[width=\linewidth]{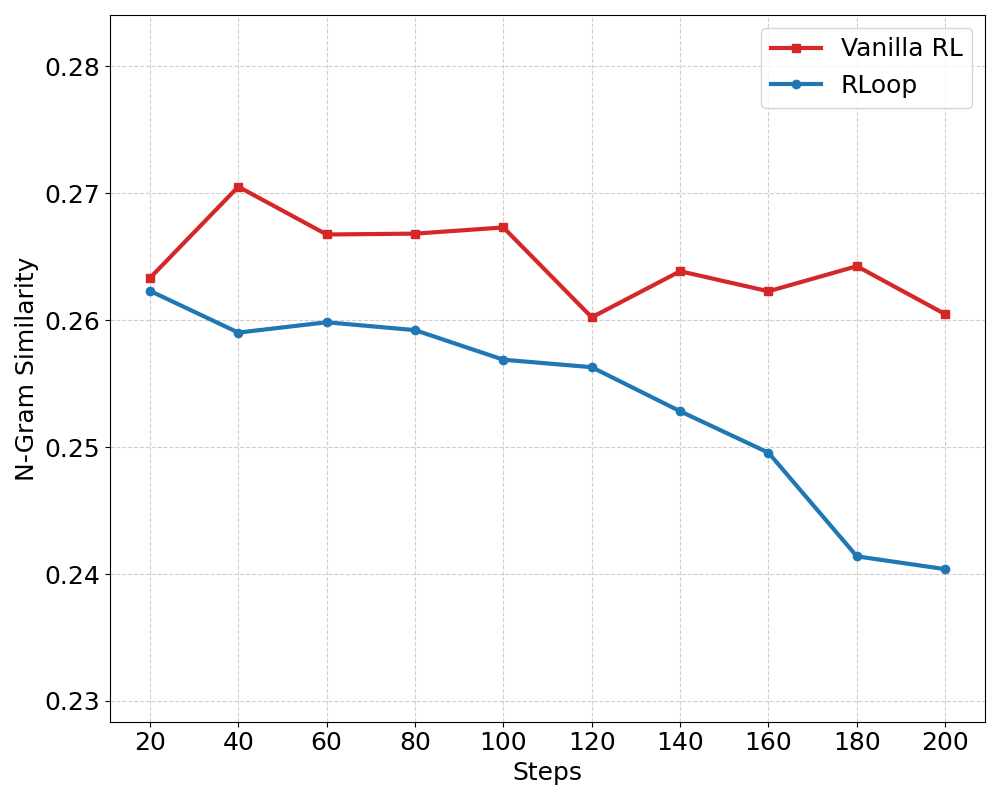}
        \caption{N-gram similarity.}
        \label{fig:diversity-comparison}
    \end{subfigure}
    \begin{subfigure}[b]{0.32 \linewidth}
        \centering
        \includegraphics[width=\linewidth]{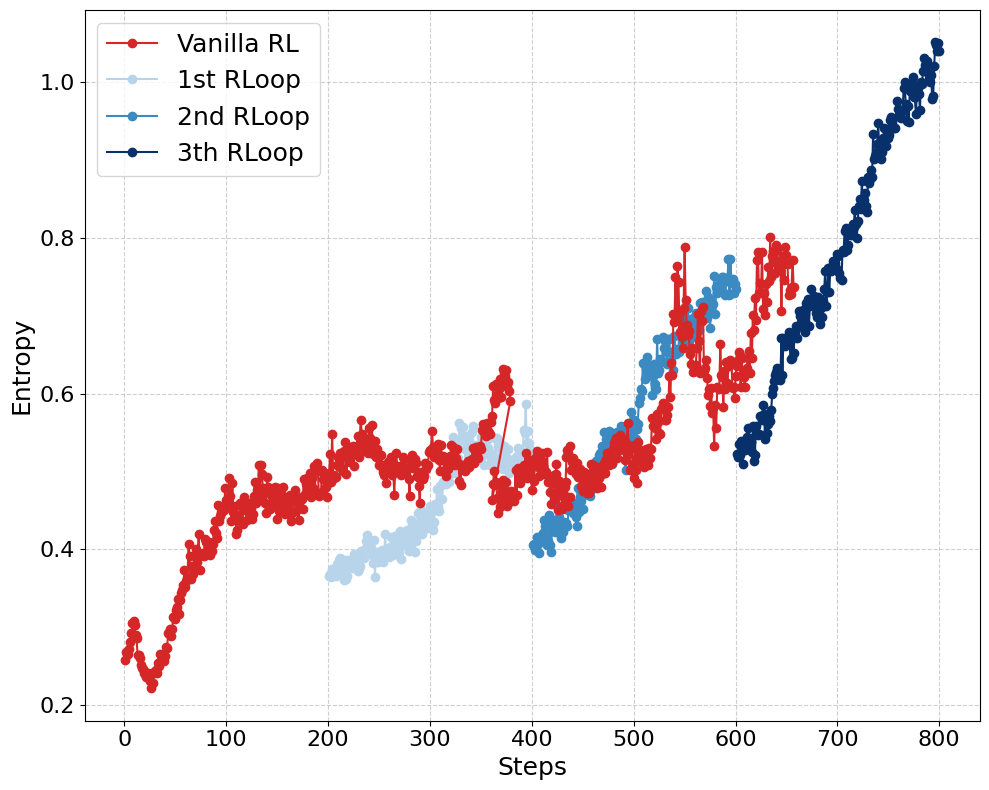}
        \caption{Entropy.}
        \label{fig:entropy-comparison}
    \end{subfigure}
    \caption{(a): Analysis of RLoop's mechanisms compared to vanilla RL. (a) Differential forgetting matrix (Vanilla RL Forgetting - RLoop Forgetting). Blue indicates RLoop forgets less. (b) N-gram similarity comparison, where lower values imply higher diversity. (c) Token-level policy entropy over training steps.}
    \label{fig:forget-and-diversity-comparison}
\end{figure*}

\subsection{Scalability Analysis}
\label{sec:scaling-reinit}
In this section, we investigate the scalability of RLoop by analyzing its performance over an extended number of iterations and comparing its learning dynamic against vanilla RL.

\paragraph{Scaling with More Iterations} 

Figure \ref{fig:scale-reinit} illustrates the performance of RLoop as a function of the number of iterations. The results demonstrate a clear positive scaling trend: performance on both accuracy (Avg@32) and pass@k metrics improves with additional iterations. This trend is particularly evident on the Omni-Math and MATH benchmarks. Notably, the improvement in pass@k scores is more pronounced than the gains in accuracy, suggesting that continued iterations primarily enhance the model's ability to generate a diverse set of correct solutions.

\paragraph{Contrasting Scalability Dynamics} 
To understand how RLoop utilizes computational budget differently from vanilla RL, we plot its performance against vanilla RL on a continuous training step axis. As shown in Figure \ref{fig:compare-reinit-rl-acc}, each 200-step RL phase of a RLoop iteration is juxtaposed with the corresponding training window of the vanilla RL baseline.

The comparison reveals a stark contrast. Vanilla RL (the red curve) exhibits classic overfitting: its performance on the validation set peaks around 300 steps and then steadily degrades, indicating that further training is detrimental. In contrast, RLoop effectively leverages the additional computational budget. While vanilla RL's performance deteriorates, RLoop continues to achieve new performance heights with each subsequent iteration.

A closer examination reveals a fascinating dynamic within each RLoop iteration. The performance often rises before plateauing or slightly fluctuating, mirroring the overfitting pattern of vanilla RL on a micro-scale. However, the crucial difference is that each new iteration begins from a superior starting point established by the RFT phase. This cyclical process allows RLoop to progressively climb to higher performance levels, escaping the terminal decline that plagues standard RL.

\subsection{Why can RLoop Improve Generalization of RL?}
Having established RLoop's superior performance in Table \ref{tab:main-results} and Figure \ref{fig:scale-reinit}, we now dissect the underlying mechanisms responsible for its improved generalization. Our analysis focuses on three key areas: \textbf{catastrophic forgetting}, \textbf{trajectory diversity}, and \textbf{policy entropy}. To facilitate a direct comparison, we adopt the experimental setup from Section \ref{sec:scaling-reinit}, aligning the $i$-th iteration of RLoop with the corresponding training window of the vanilla RL baseline (steps $200i$ to $200(i+1)$).

\paragraph{Less Forgetting}
To quantify the difference in knowledge retention, we compute a differential forgetting matrix, defined as the forgetting rate of vanilla RL minus that of RLoop. As shown in Figure \ref{fig:forget-comparison}, blue cells indicate that RLoop forgets less (a positive outcome), while red cells indicate the opposite. The matrix is predominantly blue, providing strong visual evidence that RLoop generally suffers from less catastrophic forgetting than the standard RL baseline.

A deeper analysis reveals a crucial distinction between \textit{intra-iteration} (within an RL phase) and \textit{inter-iteration} (across RFT resets) forgetting. The forgetting rates within each 200-step RL phase of RLoop (the block-diagonal regions) are comparable to those of vanilla RL, exhibiting a similar level of instability. However, the forgetting between iterations (the off-diagonal blocks) is substantially lower for RLoop. This indicates that the RFT phase is highly effective at consolidating knowledge and serving as a stable anchor, preventing the long-term, catastrophic forgetting that plagues uninterrupted RL training.

\paragraph{Better Trajectory Diversity}
We next examine trajectory diversity by comparing the n-gram similarity of generated solutions, a metric inversely proportional to diversity. The process of estimating n-gram similarity is shown in Appendix \ref{app:similarity-calculation}. Figure \ref{fig:diversity-comparison} shows that RLoop consistently maintains a lower average n-gram similarity than vanilla RL throughout the training process. Since lower similarity corresponds to higher diversity, this result demonstrates that RLoop promotes a more diverse set of generated solutions. This heightened diversity is a key factor contributing to RLoop's superior generalization and, in particular, its significantly improved pass@k performance.

\paragraph{High Entropy}
Policy entropy is widely regarded as a proxy for exploration in RL \citep{beyond-passat2,entropy-rl,prolong-rl}. We therefore compare the token-level entropy of policies trained with RLoop and vanilla RL. As shown in Figure \ref{fig:entropy-comparison}, the entropy for both methods generally increases over time, and crucially, RLoop maintains a policy entropy comparable to that of vanilla RL. It suggests that RLoop's benefits of reduced forgetting and increased diversity are achieved without sacrificing policy exploration.

\subsection{RLoop Improves Training Stability}
\begin{figure*}[t]
    \centering
    \begin{subfigure}[b]{0.4\linewidth}
        \centering
        \includegraphics[width=\linewidth]{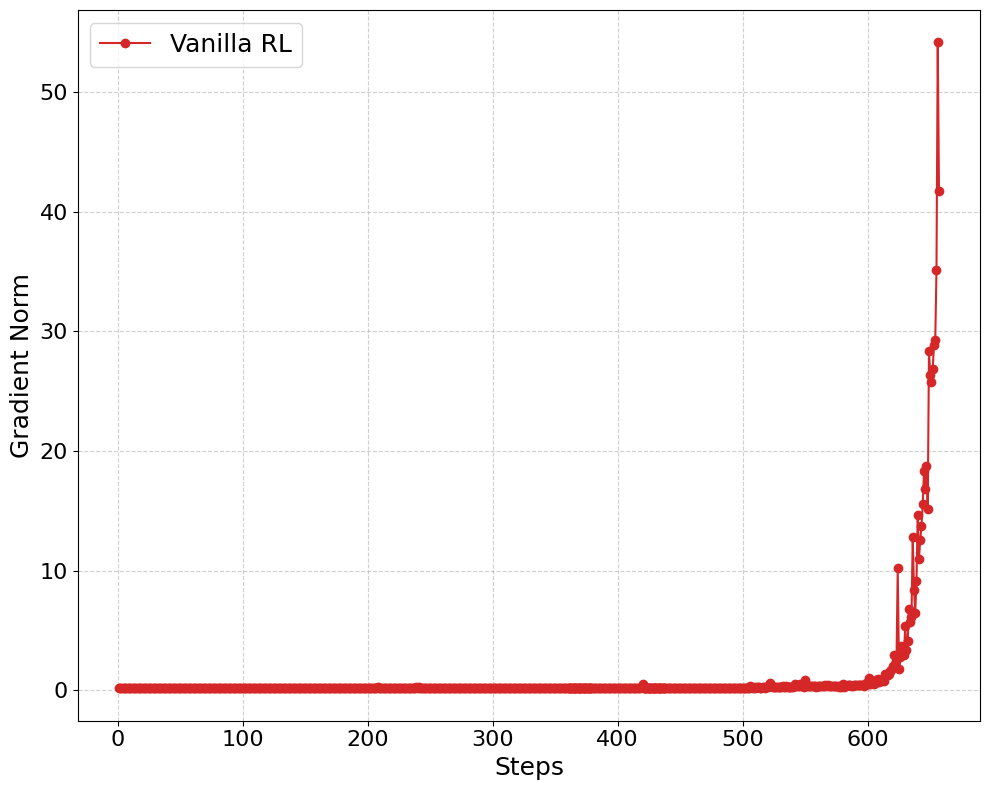}
        \caption{Gradient Norm of Vanilla RL}
        \label{fig:rl-grad-norm}
    \end{subfigure}
    \begin{subfigure}[b]{0.4\linewidth}
        \centering
        \includegraphics[width=\linewidth]{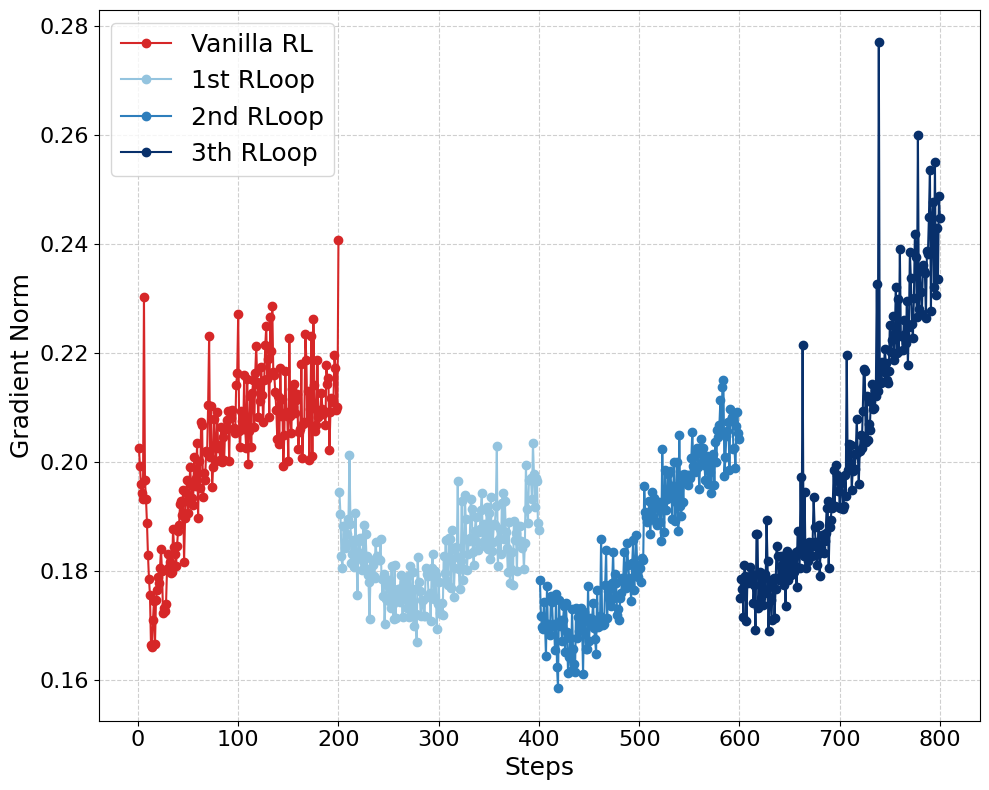}
        \caption{Gradient Norm of RLoop}
        \label{fig:reinit-grad-norm}
    \end{subfigure}
    \caption{Comparison of gradient norm stability. (a) Vanilla RL exhibits explosive gradient growth after 600 steps, leading to training collapse. (b) RLoop maintains a stable, low gradient norm throughout its iterative training process.}
    \label{fig:grad-norm-compare}
\end{figure*}

A well-documented challenge in prolonged RL fine-tuning of LLMs is training instability, often manifesting as gradient explosion and catastrophic training collapse \cite{off-policy-rl,defeating-nondeterminism,stable-moe,rl-Collapse}. Our experiments with vanilla RL confirm this issue; we observed a training collapse around 750 steps, preceded by an uncontrolled surge in the gradient norm. We find that our RLoop framework inherently mitigates this instability.

Figure \ref{fig:grad-norm-compare} provides a clear illustration of this effect. The gradient norm for the vanilla RL baseline (Figure \ref{fig:rl-grad-norm}) remains manageable for approximately 600 steps before experiencing explosive growth, quickly exceeding 50.0 and causing the training process to fail. In stark contrast, the gradient norm for RLoop (Figure \ref{fig:reinit-grad-norm}) remains remarkably stable and bounded. Even after three full iterations, corresponding to a total of 800 RL steps, the norm stays below 0.3, demonstrating the framework's robustness against the instabilities that plague standard RL.

The source of this stability lies in RLoop's cyclical ``reset'' mechanism. Instead of a single, prolonged optimization trajectory, RLoop performs a series of shorter, bounded RL explorations. Crucially, each exploration phase begins from a ``refreshed'' policy. This policy is not the potentially unstable endpoint of the previous RL phase, but rather a new model created by fine-tuning the original, stable base model on a small, high-quality dataset of expert trajectories. This periodic re-anchoring to a stable base prevents the policy from drifting into volatile regions of the parameter space. In contrast, the vanilla RL process, after 750 steps (equivalent to approximately 45 epochs over the training data), is likely over-optimizing on a fixed dataset, leading to the observed gradient explosion.

\section{Conclusion}
In this work, we identified ``RL overfitting'' as a critical challenge in RLVR, linking it to catastrophic forgetting and the under-utilization of inter-step policy diversity. To address this, we proposed RLoop, an iterative framework that transforms RL's instability into a strength by cyclically alternating between an RL exploration phase to generate diverse solutions and an RFT exploitation phase to consolidate knowledge. Our experiments demonstrate that RLoop significantly outperforms vanilla RL, particularly in pass@k metrics, by mitigating long-term forgetting, enhancing solution diversity, and ensuring training stability. By reframing training instability as a valuable source of exploration, our work offers a robust and principled solution to current RL challenges and paves the way for more stable, generalizable, and powerful reasoning models.

\bibliography{main}

@article{bridge,
  author       = {Liang Chen and
                  Xueting Han and
                  Li Shen and
                  Jing Bai and
                  Kam{-}Fai Wong},
  title        = {Beyond Two-Stage Training: Cooperative {SFT} and {RL} for {LLM} Reasoning},
  journal      = {CoRR},
  volume       = {abs/2509.06948},
  year         = {2025},
  url          = {https://doi.org/10.48550/arXiv.2509.06948},
  doi          = {10.48550/ARXIV.2509.06948},
  eprinttype    = {arXiv},
  eprint       = {2509.06948},
  bibsource    = {dblp computer science bibliography, https://dblp.org}
}

@article{ReLIFT,
  author       = {Lu Ma and
                  Hao Liang and
                  Meiyi Qiang and
                  Lexiang Tang and
                  Xiaochen Ma and
                  Zhen Hao Wong and
                  Junbo Niu and
                  Chengyu Shen and
                  Runming He and
                  Bin Cui and
                  Wentao Zhang},
  title        = {Learning What Reinforcement Learning Can't: Interleaved Online
                  Fine-Tuning for Hardest Questions},
  journal      = {CoRR},
  volume       = {abs/2506.07527},
  year         = {2025},
  url          = {https://doi.org/10.48550/arXiv.2506.07527},
  doi          = {10.48550/ARXIV.2506.07527},
  eprinttype    = {arXiv},
  eprint       = {2506.07527},
  bibsource    = {dblp computer science bibliography, https://dblp.org}
}

@article{luffy,
  author       = {Jianhao Yan and
                  Yafu Li and
                  Zican Hu and
                  Zhi Wang and
                  Ganqu Cui and
                  Xiaoye Qu and
                  Yu Cheng and
                  Yue Zhang},
  title        = {Learning to Reason under Off-Policy Guidance},
  journal      = {CoRR},
  volume       = {abs/2504.14945},
  year         = {2025},
  url          = {https://doi.org/10.48550/arXiv.2504.14945},
  doi          = {10.48550/ARXIV.2504.14945},
  eprinttype    = {arXiv},
  eprint       = {2504.14945},
  bibsource    = {dblp computer science bibliography, https://dblp.org}
}

@article{chord,
  author       = {Wenhao Zhang and
                  Yuexiang Xie and
                  Yuchang Sun and
                  Yanxi Chen and
                  Guoyin Wang and
                  Yaliang Li and
                  Bolin Ding and
                  Jingren Zhou},
  title        = {On-Policy {RL} Meets Off-Policy Experts: Harmonizing Supervised Fine-Tuning
                  and Reinforcement Learning via Dynamic Weighting},
  journal      = {CoRR},
  volume       = {abs/2508.11408},
  year         = {2025},
  url          = {https://doi.org/10.48550/arXiv.2508.11408},
  doi          = {10.48550/ARXIV.2508.11408},
  eprinttype    = {arXiv},
  eprint       = {2508.11408},
  bibsource    = {dblp computer science bibliography, https://dblp.org}
}

@misc{MIFO,
      title={Mitigating Forgetting Between Supervised and Reinforcement Learning Yields Stronger Reasoners}, 
      author={Xiangchi Yuan and Xiang Chen and Tong Yu and Dachuan Shi and Can Jin and Wenke Lee and Saayan Mitra},
      year={2025},
      eprint={2510.04454},
      archivePrefix={arXiv},
      primaryClass={cs.CL},
      url={https://arxiv.org/abs/2510.04454}, 
}

@article{limited-rl,
  author       = {Yang Yue and
                  Zhiqi Chen and
                  Rui Lu and
                  Andrew Zhao and
                  Zhaokai Wang and
                  Yang Yue and
                  Shiji Song and
                  Gao Huang},
  title        = {Does Reinforcement Learning Really Incentivize Reasoning Capacity
                  in LLMs Beyond the Base Model?},
  journal      = {CoRR},
  volume       = {abs/2504.13837},
  year         = {2025},
  url          = {https://doi.org/10.48550/arXiv.2504.13837},
  doi          = {10.48550/ARXIV.2504.13837},
  eprinttype    = {arXiv},
  eprint       = {2504.13837},
  bibsource    = {dblp computer science bibliography, https://dblp.org}
}

@article{prolong-rl,
  author       = {Mingjie Liu and
                  Shizhe Diao and
                  Ximing Lu and
                  Jian Hu and
                  Xin Dong and
                  Yejin Choi and
                  Jan Kautz and
                  Yi Dong},
  title        = {ProRL: Prolonged Reinforcement Learning Expands Reasoning Boundaries
                  in Large Language Models},
  journal      = {CoRR},
  volume       = {abs/2505.24864},
  year         = {2025},
  url          = {https://doi.org/10.48550/arXiv.2505.24864},
  doi          = {10.48550/ARXIV.2505.24864},
  eprinttype    = {arXiv},
  eprint       = {2505.24864},
  bibsource    = {dblp computer science bibliography, https://dblp.org}
}

@article{QuestA,
  author       = {Jiazheng Li and
                  Hong Lu and
                  Kaiyue Wen and
                  Zaiwen Yang and
                  Jiaxuan Gao and
                  Hongzhou Lin and
                  Yi Wu and
                  Jingzhao Zhang},
  title        = {QuestA: Expanding Reasoning Capacity in LLMs via Question Augmentation},
  journal      = {CoRR},
  volume       = {abs/2507.13266},
  year         = {2025},
  url          = {https://doi.org/10.48550/arXiv.2507.13266},
  doi          = {10.48550/ARXIV.2507.13266},
  eprinttype    = {arXiv},
  eprint       = {2507.13266},
  bibsource    = {dblp computer science bibliography, https://dblp.org}
}

@article{beyond-passat1,
  author       = {Xiao Liang and
                  Zhongzhi Li and
                  Yeyun Gong and
                  Yelong Shen and
                  Ying Nian Wu and
                  Zhijiang Guo and
                  Weizhu Chen},
  title        = {Beyond Pass@1: Self-Play with Variational Problem Synthesis Sustains
                  {RLVR}},
  journal      = {CoRR},
  volume       = {abs/2508.14029},
  year         = {2025},
  url          = {https://doi.org/10.48550/arXiv.2508.14029},
  doi          = {10.48550/ARXIV.2508.14029},
  eprinttype    = {arXiv},
  eprint       = {2508.14029},
  bibsource    = {dblp computer science bibliography, https://dblp.org}
}

@misc{zero-reward,
title={What Can You Do When You Have Zero Rewards During RL?},
author={Jatin Prakash and Anirudh Buvanesh},
year={2025},
howpublished={\url{https://spiffy-airbus-472.notion.site/What-Can-You-Do-When-You-Have-Zero-Rewards-During-RL-260429bdb7308024b6bdd3ed4f64c15f}},
note={Notion Blog},
}

@misc{knapsack-rl,
      title={Knapsack RL: Unlocking Exploration of LLMs via Optimizing Budget Allocation}, 
      author={Ziniu Li and Congliang Chen and Tianyun Yang and Tian Ding and Ruoyu Sun and Ge Zhang and Wenhao Huang and Zhi-Quan Luo},
      year={2025},
      eprint={2509.25849},
      archivePrefix={arXiv},
      primaryClass={cs.LG},
      url={https://arxiv.org/abs/2509.25849}, 
}

@article{entropy-rl,
  author       = {Ganqu Cui and
                  Yuchen Zhang and
                  Jiacheng Chen and
                  Lifan Yuan and
                  Zhi Wang and
                  Yuxin Zuo and
                  Haozhan Li and
                  Yuchen Fan and
                  Huayu Chen and
                  Weize Chen and
                  Zhiyuan Liu and
                  Hao Peng and
                  Lei Bai and
                  Wanli Ouyang and
                  Yu Cheng and
                  Bowen Zhou and
                  Ning Ding},
  title        = {The Entropy Mechanism of Reinforcement Learning for Reasoning Language
                  Models},
  journal      = {CoRR},
  volume       = {abs/2505.22617},
  year         = {2025},
  url          = {https://doi.org/10.48550/arXiv.2505.22617},
  doi          = {10.48550/ARXIV.2505.22617},
  eprinttype    = {arXiv},
  eprint       = {2505.22617},
  bibsource    = {dblp computer science bibliography, https://dblp.org}
}

@article{beyond-passat2,
  author       = {Shenzhi Wang and
                  Le Yu and
                  Chang Gao and
                  Chujie Zheng and
                  Shixuan Liu and
                  Rui Lu and
                  Kai Dang and
                  Xionghui Chen and
                  Jianxin Yang and
                  Zhenru Zhang and
                  Yuqiong Liu and
                  An Yang and
                  Andrew Zhao and
                  Yang Yue and
                  Shiji Song and
                  Bowen Yu and
                  Gao Huang and
                  Junyang Lin},
  title        = {Beyond the 80/20 Rule: High-Entropy Minority Tokens Drive Effective
                  Reinforcement Learning for {LLM} Reasoning},
  journal      = {CoRR},
  volume       = {abs/2506.01939},
  year         = {2025},
  url          = {https://doi.org/10.48550/arXiv.2506.01939},
  doi          = {10.48550/ARXIV.2506.01939},
  eprinttype    = {arXiv},
  eprint       = {2506.01939},
  bibsource    = {dblp computer science bibliography, https://dblp.org}
}

@article{one-shot-rl,
  author       = {Yiping Wang and
                  Qing Yang and
                  Zhiyuan Zeng and
                  Liliang Ren and
                  Lucas Liu and
                  Baolin Peng and
                  Hao Cheng and
                  Xuehai He and
                  Kuan Wang and
                  Jianfeng Gao and
                  Weizhu Chen and
                  Shuohang Wang and
                  Simon Shaolei Du and
                  Yelong Shen},
  title        = {Reinforcement Learning for Reasoning in Large Language Models with
                  One Training Example},
  journal      = {CoRR},
  volume       = {abs/2504.20571},
  year         = {2025},
  url          = {https://doi.org/10.48550/arXiv.2504.20571},
  doi          = {10.48550/ARXIV.2504.20571},
  eprinttype    = {arXiv},
  eprint       = {2504.20571},
  bibsource    = {dblp computer science bibliography, https://dblp.org}
}

@inproceedings{
causal-rl,
title={Generalization of {RLVR} Using Causal Reasoning as a Testbed},
author={Anonymous},
booktitle={Submitted to The Fourteenth International Conference on Learning Representations},
year={2025},
url={https://openreview.net/forum?id=DZjbL9BuHs},
note={under review}
}

@article{med-rlvr,
  author       = {Sheng Zhang and
                  Qianchu Liu and
                  Guanghui Qin and
                  Tristan Naumann and
                  Hoifung Poon},
  title        = {Med-RLVR: Emerging Medical Reasoning from a 3B base model via reinforcement
                  Learning},
  journal      = {CoRR},
  volume       = {abs/2502.19655},
  year         = {2025},
  url          = {https://doi.org/10.48550/arXiv.2502.19655},
  doi          = {10.48550/ARXIV.2502.19655},
  eprinttype    = {arXiv},
  eprint       = {2502.19655},
  bibsource    = {dblp computer science bibliography, https://dblp.org}
}

@article{sft-rl-generalizes,
  author       = {Tianzhe Chu and
                  Yuexiang Zhai and
                  Jihan Yang and
                  Shengbang Tong and
                  Saining Xie and
                  Dale Schuurmans and
                  Quoc V. Le and
                  Sergey Levine and
                  Yi Ma},
  title        = {{SFT} Memorizes, {RL} Generalizes: {A} Comparative Study of Foundation
                  Model Post-training},
  journal      = {CoRR},
  volume       = {abs/2501.17161},
  year         = {2025},
  url          = {https://doi.org/10.48550/arXiv.2501.17161},
  doi          = {10.48550/ARXIV.2501.17161},
  eprinttype    = {arXiv},
  eprint       = {2501.17161},
  bibsource    = {dblp computer science bibliography, https://dblp.org}
}

@article{deepseek-r1,
  author       = {DeepSeek{-}AI and
                  Daya Guo and
                  Dejian Yang and
                  Haowei Zhang and
                  Junxiao Song and
                  Ruoyu Zhang and
                  Runxin Xu and
                  Qihao Zhu and
                  Shirong Ma and
                  Peiyi Wang and
                  Xiao Bi and
                  Xiaokang Zhang and
                  Xingkai Yu and
                  Yu Wu and
                  Z. F. Wu and
                  Zhibin Gou and
                  Zhihong Shao and
                  Zhuoshu Li and
                  Ziyi Gao and
                  Aixin Liu and
                  Bing Xue and
                  Bingxuan Wang and
                  Bochao Wu and
                  Bei Feng and
                  Chengda Lu and
                  Chenggang Zhao and
                  Chengqi Deng and
                  Chenyu Zhang and
                  Chong Ruan and
                  Damai Dai and
                  Deli Chen and
                  Dongjie Ji and
                  Erhang Li and
                  Fangyun Lin and
                  Fucong Dai and
                  Fuli Luo and
                  Guangbo Hao and
                  Guanting Chen and
                  Guowei Li and
                  H. Zhang and
                  Han Bao and
                  Hanwei Xu and
                  Haocheng Wang and
                  Honghui Ding and
                  Huajian Xin and
                  Huazuo Gao and
                  Hui Qu and
                  Hui Li and
                  Jianzhong Guo and
                  Jiashi Li and
                  Jiawei Wang and
                  Jingchang Chen and
                  Jingyang Yuan and
                  Junjie Qiu and
                  Junlong Li and
                  J. L. Cai and
                  Jiaqi Ni and
                  Jian Liang and
                  Jin Chen and
                  Kai Dong and
                  Kai Hu and
                  Kaige Gao and
                  Kang Guan and
                  Kexin Huang and
                  Kuai Yu and
                  Lean Wang and
                  Lecong Zhang and
                  Liang Zhao and
                  Litong Wang and
                  Liyue Zhang and
                  Lei Xu and
                  Leyi Xia and
                  Mingchuan Zhang and
                  Minghua Zhang and
                  Minghui Tang and
                  Meng Li and
                  Miaojun Wang and
                  Mingming Li and
                  Ning Tian and
                  Panpan Huang and
                  Peng Zhang and
                  Qiancheng Wang and
                  Qinyu Chen and
                  Qiushi Du and
                  Ruiqi Ge and
                  Ruisong Zhang and
                  Ruizhe Pan and
                  Runji Wang and
                  R. J. Chen and
                  R. L. Jin and
                  Ruyi Chen and
                  Shanghao Lu and
                  Shangyan Zhou and
                  Shanhuang Chen and
                  Shengfeng Ye and
                  Shiyu Wang and
                  Shuiping Yu and
                  Shunfeng Zhou and
                  Shuting Pan and
                  S. S. Li},
  title        = {DeepSeek-R1: Incentivizing Reasoning Capability in LLMs via Reinforcement
                  Learning},
  journal      = {CoRR},
  volume       = {abs/2501.12948},
  year         = {2025},
  url          = {https://doi.org/10.48550/arXiv.2501.12948},
  doi          = {10.48550/ARXIV.2501.12948},
  eprinttype    = {arXiv},
  eprint       = {2501.12948},
  bibsource    = {dblp computer science bibliography, https://dblp.org}
}

@misc{vae,
      title={Auto-Encoding Variational Bayes}, 
      author={Diederik P Kingma and Max Welling},
      year={2022},
      eprint={1312.6114},
      archivePrefix={arXiv},
      primaryClass={stat.ML},
      url={https://arxiv.org/abs/1312.6114}, 
}

@inproceedings{instruct-gpt,
  author       = {Long Ouyang and
                  Jeffrey Wu and
                  Xu Jiang and
                  Diogo Almeida and
                  Carroll L. Wainwright and
                  Pamela Mishkin and
                  Chong Zhang and
                  Sandhini Agarwal and
                  Katarina Slama and
                  Alex Ray and
                  John Schulman and
                  Jacob Hilton and
                  Fraser Kelton and
                  Luke Miller and
                  Maddie Simens and
                  Amanda Askell and
                  Peter Welinder and
                  Paul F. Christiano and
                  Jan Leike and
                  Ryan Lowe},
  editor       = {Sanmi Koyejo and
                  S. Mohamed and
                  A. Agarwal and
                  Danielle Belgrave and
                  K. Cho and
                  A. Oh},
  title        = {Training language models to follow instructions with human feedback},
  booktitle    = {Advances in Neural Information Processing Systems 35: Annual Conference
                  on Neural Information Processing Systems 2022, NeurIPS 2022, New Orleans,
                  LA, USA, November 28 - December 9, 2022},
  year         = {2022},
  url          = {http://papers.nips.cc/paper\_files/paper/2022/hash/b1efde53be364a73914f58805a001731-Abstract-Conference.html},
  bibsource    = {dblp computer science bibliography, https://dblp.org}
}

@article{openthoughts,
  author       = {Etash Kumar Guha and
                  Ryan Marten and
                  Sedrick Keh and
                  Negin Raoof and
                  Georgios Smyrnis and
                  Hritik Bansal and
                  Marianna Nezhurina and
                  Jean Mercat and
                  Trung Vu and
                  Zayne Sprague and
                  Ashima Suvarna and
                  Benjamin Feuer and
                  Liangyu Chen and
                  Zaid Khan and
                  Eric Frankel and
                  Sachin Grover and
                  Caroline Choi and
                  Niklas Muennighoff and
                  Shiye Su and
                  Wanjia Zhao and
                  John Yang and
                  Shreyas Pimpalgaonkar and
                  Kartik Sharma and
                  Charlie Cheng{-}Jie Ji and
                  Yichuan Deng and
                  Sarah M. Pratt and
                  Vivek Ramanujan and
                  Jon Saad{-}Falcon and
                  Jeffrey Li and
                  Achal Dave and
                  Alon Albalak and
                  Kushal Arora and
                  Blake Wulfe and
                  Chinmay Hegde and
                  Greg Durrett and
                  Sewoong Oh and
                  Mohit Bansal and
                  Saadia Gabriel and
                  Aditya Grover and
                  Kai{-}Wei Chang and
                  Vaishaal Shankar and
                  Aaron Gokaslan and
                  Mike A. Merrill and
                  Tatsunori Hashimoto and
                  Yejin Choi and
                  Jenia Jitsev and
                  Reinhard Heckel and
                  Maheswaran Sathiamoorthy and
                  Alexandros G. Dimakis and
                  Ludwig Schmidt},
  title        = {OpenThoughts: Data Recipes for Reasoning Models},
  journal      = {CoRR},
  volume       = {abs/2506.04178},
  year         = {2025},
  url          = {https://doi.org/10.48550/arXiv.2506.04178},
  doi          = {10.48550/ARXIV.2506.04178},
  eprinttype    = {arXiv},
  eprint       = {2506.04178},
  bibsource    = {dblp computer science bibliography, https://dblp.org}
}

@article{OctoThinker,
  author       = {Zengzhi Wang and
                  Fan Zhou and
                  Xuefeng Li and
                  Pengfei Liu},
  title        = {OctoThinker: Mid-training Incentivizes Reinforcement Learning Scaling},
  journal      = {CoRR},
  volume       = {abs/2506.20512},
  year         = {2025},
  url          = {https://doi.org/10.48550/arXiv.2506.20512},
  doi          = {10.48550/ARXIV.2506.20512},
  eprinttype    = {arXiv},
  eprint       = {2506.20512},
  bibsource    = {dblp computer science bibliography, https://dblp.org}
}

@article{BRiTE,
  author       = {Han Zhong and
                  Yutong Yin and
                  Shenao Zhang and
                  Xiaojun Xu and
                  Yuanxin Liu and
                  Yifei Zuo and
                  Zhihan Liu and
                  Boyi Liu and
                  Sirui Zheng and
                  Hongyi Guo and
                  Liwei Wang and
                  Mingyi Hong and
                  Zhaoran Wang},
  title        = {BRiTE: Bootstrapping Reinforced Thinking Process to Enhance Language
                  Model Reasoning},
  journal      = {CoRR},
  volume       = {abs/2501.18858},
  year         = {2025},
  url          = {https://doi.org/10.48550/arXiv.2501.18858},
  doi          = {10.48550/ARXIV.2501.18858},
  eprinttype    = {arXiv},
  eprint       = {2501.18858},
  bibsource    = {dblp computer science bibliography, https://dblp.org}
}

@article{self-rewarding,
  author       = {Haolin Chen and
                  Yihao Feng and
                  Zuxin Liu and
                  Weiran Yao and
                  Akshara Prabhakar and
                  Shelby Heinecke and
                  Ricky Ho and
                  Phil Mui and
                  Silvio Savarese and
                  Caiming Xiong and
                  Huan Wang},
  title        = {Language Models are Hidden Reasoners: Unlocking Latent Reasoning Capabilities
                  via Self-Rewarding},
  journal      = {CoRR},
  volume       = {abs/2411.04282},
  year         = {2024},
  url          = {https://doi.org/10.48550/arXiv.2411.04282},
  doi          = {10.48550/ARXIV.2411.04282},
  eprinttype    = {arXiv},
  eprint       = {2411.04282},
  bibsource    = {dblp computer science bibliography, https://dblp.org}
}

@article{rft,
  author       = {Zheng Yuan and
                  Hongyi Yuan and
                  Chengpeng Li and
                  Guanting Dong and
                  Chuanqi Tan and
                  Chang Zhou},
  title        = {Scaling Relationship on Learning Mathematical Reasoning with Large
                  Language Models},
  journal      = {CoRR},
  volume       = {abs/2308.01825},
  year         = {2023},
  url          = {https://doi.org/10.48550/arXiv.2308.01825},
  doi          = {10.48550/ARXIV.2308.01825},
  eprinttype    = {arXiv},
  eprint       = {2308.01825},
  bibsource    = {dblp computer science bibliography, https://dblp.org}
}

@article{DAPO,
  author       = {Qiying Yu and
                  Zheng Zhang and
                  Ruofei Zhu and
                  Yufeng Yuan and
                  Xiaochen Zuo and
                  Yu Yue and
                  Tiantian Fan and
                  Gaohong Liu and
                  Lingjun Liu and
                  Xin Liu and
                  Haibin Lin and
                  Zhiqi Lin and
                  Bole Ma and
                  Guangming Sheng and
                  Yuxuan Tong and
                  Chi Zhang and
                  Mofan Zhang and
                  Wang Zhang and
                  Hang Zhu and
                  Jinhua Zhu and
                  Jiaze Chen and
                  Jiangjie Chen and
                  Chengyi Wang and
                  Hongli Yu and
                  Weinan Dai and
                  Yuxuan Song and
                  Xiangpeng Wei and
                  Hao Zhou and
                  Jingjing Liu and
                  Wei{-}Ying Ma and
                  Ya{-}Qin Zhang and
                  Lin Yan and
                  Mu Qiao and
                  Yonghui Wu and
                  Mingxuan Wang},
  title        = {{DAPO:} An Open-Source {LLM} Reinforcement Learning System at Scale},
  journal      = {CoRR},
  volume       = {abs/2503.14476},
  year         = {2025},
  url          = {https://doi.org/10.48550/arXiv.2503.14476},
  doi          = {10.48550/ARXIV.2503.14476},
  eprinttype    = {arXiv},
  eprint       = {2503.14476},
  bibsource    = {dblp computer science bibliography, https://dblp.org}
}

@inproceedings{minervamath,
  author       = {Aitor Lewkowycz and
                  Anders Andreassen and
                  David Dohan and
                  Ethan Dyer and
                  Henryk Michalewski and
                  Vinay V. Ramasesh and
                  Ambrose Slone and
                  Cem Anil and
                  Imanol Schlag and
                  Theo Gutman{-}Solo and
                  Yuhuai Wu and
                  Behnam Neyshabur and
                  Guy Gur{-}Ari and
                  Vedant Misra},
  editor       = {Sanmi Koyejo and
                  S. Mohamed and
                  A. Agarwal and
                  Danielle Belgrave and
                  K. Cho and
                  A. Oh},
  title        = {Solving Quantitative Reasoning Problems with Language Models},
  booktitle    = {Advances in Neural Information Processing Systems 35: Annual Conference
                  on Neural Information Processing Systems 2022, NeurIPS 2022, New Orleans,
                  LA, USA, November 28 - December 9, 2022},
  year         = {2022},
  url          = {http://papers.nips.cc/paper\_files/paper/2022/hash/18abbeef8cfe9203fdf9053c9c4fe191-Abstract-Conference.html},
  bibsource    = {dblp computer science bibliography, https://dblp.org}
}

@inproceedings{math-dataset,
  author       = {Dan Hendrycks and
                  Collin Burns and
                  Saurav Kadavath and
                  Akul Arora and
                  Steven Basart and
                  Eric Tang and
                  Dawn Song and
                  Jacob Steinhardt},
  editor       = {Joaquin Vanschoren and
                  Sai{-}Kit Yeung},
  title        = {Measuring Mathematical Problem Solving With the {MATH} Dataset},
  booktitle    = {Proceedings of the Neural Information Processing Systems Track on
                  Datasets and Benchmarks 1, NeurIPS Datasets and Benchmarks 2021, December
                  2021, virtual},
  year         = {2021},
  url          = {https://datasets-benchmarks-proceedings.neurips.cc/paper/2021/hash/be83ab3ecd0db773eb2dc1b0a17836a1-Abstract-round2.html},
  bibsource    = {dblp computer science bibliography, https://dblp.org}
}

@inproceedings{omini-math,
  author       = {Bofei Gao and
                  Feifan Song and
                  Zhe Yang and
                  Zefan Cai and
                  Yibo Miao and
                  Qingxiu Dong and
                  Lei Li and
                  Chenghao Ma and
                  Liang Chen and
                  Runxin Xu and
                  Zhengyang Tang and
                  Benyou Wang and
                  Daoguang Zan and
                  Shanghaoran Quan and
                  Ge Zhang and
                  Lei Sha and
                  Yichang Zhang and
                  Xuancheng Ren and
                  Tianyu Liu and
                  Baobao Chang},
  title        = {Omni-MATH: {A} Universal Olympiad Level Mathematic Benchmark for Large
                  Language Models},
  booktitle    = {The Thirteenth International Conference on Learning Representations,
                  {ICLR} 2025, Singapore, April 24-28, 2025},
  publisher    = {OpenReview.net},
  year         = {2025},
  url          = {https://openreview.net/forum?id=yaqPf0KAlN},
  bibsource    = {dblp computer science bibliography, https://dblp.org}
}

@misc{stable-moe,
      title={Stabilizing MoE Reinforcement Learning by Aligning Training and Inference Routers}, 
      author={Wenhan Ma and Hailin Zhang and Liang Zhao and Yifan Song and Yudong Wang and Zhifang Sui and Fuli Luo},
      year={2025},
      eprint={2510.11370},
      archivePrefix={arXiv},
      primaryClass={cs.CL},
      url={https://arxiv.org/abs/2510.11370}, 
}

@misc{rl-Collapse,
  title = {When Speed Kills Stability: Demystifying RL Collapse from the Inference-Training Mismatch},
  url = {https://yingru.notion.site/When-Speed-Kills-Stability-Demystifying-RL-Collapse-from-the-Inference-Training-Mismatch-271211a558b7808d8b12d403fd15edda},
  author = {Jiacai Liu and Yingru Li and Yuqian Fu and Jiawei Wang and Qian Liu and Yu Shen},
  year = {2025},
  month = september,
}

@misc{off-policy-rl,
  title = {Your Efficient RL Framework Secretly Brings You Off-Policy RL Training},
  url = {https://fengyao.notion.site/off-policy-rl},
  author = {Yao, Feng and Liu, Liyuan and Zhang, Dinghuai and Dong, Chengyu and Shang, Jingbo and Gao, Jianfeng},
  journal = {Feng Yao's Notion},
  year = {2025},
  month = aug,
}

@article{defeating-nondeterminism,
  author = {Horace He and Thinking Machines Lab},
  title = {Defeating Nondeterminism in LLM Inference},
  journal = {Thinking Machines Lab: Connectionism},
  year = {2025},
  note = {https://thinkingmachines.ai/blog/defeating-nondeterminism-in-llm-inference/},
  doi = {10.64434/tml.20250910}
}
\bibliographystyle{icml2025}

\newpage
\appendix
\appendix
\section{Detailed Calculation of Trajectory Similarity}
\label{app:similarity-calculation}

This appendix provides the detailed methodology for computing the n-gram-based trajectory similarity score, as referenced in Section \ref{sec:prelimary-study}.

To quantify the lexical consistency between different reasoning steps, we compute an n-gram-based similarity score. This metric evaluates the textual overlap between the set of solutions generated at two distinct steps, denoted as step $i$ and step $j$. For each input prompt, our model generates $N$ unique solutions. Consequently, for a single prompt, we have two sets of solutions: $S_i = \{s_{i,1}, s_{i,2}, \dots, s_{i,N}\}$ for step $i$, and $S_j = \{s_{j,1}, s_{j,2}, \dots, s_{j,N}\}$ for step $j$.

The calculation proceeds in two main stages:

\paragraph{1. Pairwise Solution Similarity via Jaccard Index} First, we define the similarity between any pair of individual solutions, one from each step ($s_{i,a} \in S_i$ and $s_{j,b} \in S_j$). This is based on their shared n-grams (we use bigrams, $n=2$, in our implementation). For each solution $s$, we generate a set of its n-grams, denoted as $G_n(s)$. The similarity between two solutions is then calculated using the Jaccard similarity coefficient:

$$
J(s_{i,a}, s_{j,b}) = \frac{|G_n(s_{i,a}) \cap G_n(s_{j,b})|}{|G_n(s_{i,a}) \cup G_n(s_{j,b})|}
$$

This value measures the proportion of shared n-grams relative to the total unique n-grams across both solutions.

\paragraph{2. Overall Step Similarity via Averaging} To obtain a single similarity score for a given prompt, we compute the Jaccard similarity for all $N \times N$ possible pairs of solutions between step $i$ and step $j$. The final similarity for that prompt, $\text{Sim}_{\text{prompt}}(S_i, S_j)$, is the arithmetic mean of these $N^2$ pairwise scores:

$$
\text{Sim}_{\text{prompt}}(S_i, S_j) = \frac{1}{N^2} \sum_{a=1}^{N} \sum_{b=1}^{N} J(s_{i,a}, s_{j,b})
$$

This aggregation provides a robust measure of the overall similarity between the two sets of solutions. The final reported similarity between step $i$ and step $j$ is the average of these $\text{Sim}_{\text{prompt}}$ scores across all prompts in the dataset.

\end{document}